\pdfoutput=1 
\PassOptionsToPackage{table}{xcolor}
\documentclass[acmtog]{acmart}
\usepackage{graphicx} 
\usepackage{enumitem} 
\newcommand{\degree}{$^\circ$}
\usepackage{booktabs}
\usepackage{float}
\usepackage{stfloats}

\newcommand{\ie}{\textit{i.e.}}

\newcommand{\ours}{StudioRecon}
\DeclareMathOperator*{\argmin}{arg\,min}

\setcopyright{cc}
\setcctype[4.0]{by-nc-nd}
\copyrightyear{2026}
\acmYear{2026}
\acmDOI{10.1145/3799902.3811165}
\acmISBN{979-8-4007-2554-8/26/07}
\acmConference[SIGGRAPH Conference Papers '26]{Special Interest Group on Computer Graphics and Interactive Techniques Conference Conference Papers}{July 19--23, 2026}{Los Angeles, CA, USA}
\acmBooktitle{Special Interest Group on Computer Graphics and Interactive Techniques Conference Conference Papers (SIGGRAPH Conference Papers '26), July 19--23, 2026, Los Angeles, CA, USA}

\begin{document}

\title{4D Human-Scene Reconstruction from Low-Overlap Captures}

\author{Minhyuk Hwang}
\authornote{Both authors contributed equally.}
\orcid{0009-0009-6026-7771}
\affiliation{%
  \institution{Seoul National University}
  \city{Seoul}
  \country{Republic of Korea}
}
\email{mhhlego@snu.ac.kr}

\author{Sangmin Kim}
\authornotemark[1]
\orcid{0009-0003-7018-1724}
\affiliation{%
  \institution{Seoul National University}
  \city{Seoul}
  \country{Republic of Korea}
}
\email{sm.kim@snu.ac.kr}

\author{Seunguk Do}
\orcid{0009-0006-3151-7338}
\affiliation{%
  \institution{Seoul National University}
  \city{Seoul}
  \country{Republic of Korea}
}
\email{seunguk.do@snu.ac.kr}

\author{Daneul Kim}
\orcid{0000-0003-2223-8063}
\affiliation{%
  \institution{Seoul National University}
  \city{Seoul}
  \country{Republic of Korea}
}
\email{carpedkm@snu.ac.kr}

\author{Jaesik Park}
\authornote{Corresponding author.}
\orcid{0000-0001-5541-409X}
\affiliation{%
  \institution{Seoul National University}
  \city{Seoul}
  \country{Republic of Korea}
}
\email{jaesik.park@snu.ac.kr}

\renewcommand{\shortauthors}{Hwang et al.}

\begin{abstract}
Existing volumetric capture of dynamic human performance achieves high fidelity with dense camera arrays. However, in real-world scenarios, only a handful of low-overlap cameras are available, which degrades the output quality and leaves large areas unobserved.
Recent 4D reconstruction methods have focused on low-overlap settings, yet they still produce noticeable artifacts in under-observed regions.
Video diffusion models have emerged as another option, but they show geometrically inconsistent results for humans.
To address these limitations, we propose \ours{}, a pipeline that reconstructs 4D human scenes from sparse, low-overlap cameras by decoupling background and humans.
We densify background supervision by synthesizing hundreds of camera-controlled novel views with a video diffusion model. We also robustly initialize deformable Gaussian humans with cross-view identity association and triangulated multi-view keypoint fitting.
Finally, our recursive enhancement module with motion-adaptive consistency injection harmonizes the composed output, thereby further avoiding remaining artifacts.
We achieve state-of-the-art novel view synthesis across four real-world datasets and demonstrate applications such as novel trajectory rendering and human replacement. Project page: \url{https://sisyphm.github.io/studiorecon-page/}.
\end{abstract}


\begin{teaserfigure}
  \centering
  \includegraphics[width=0.91\textwidth]{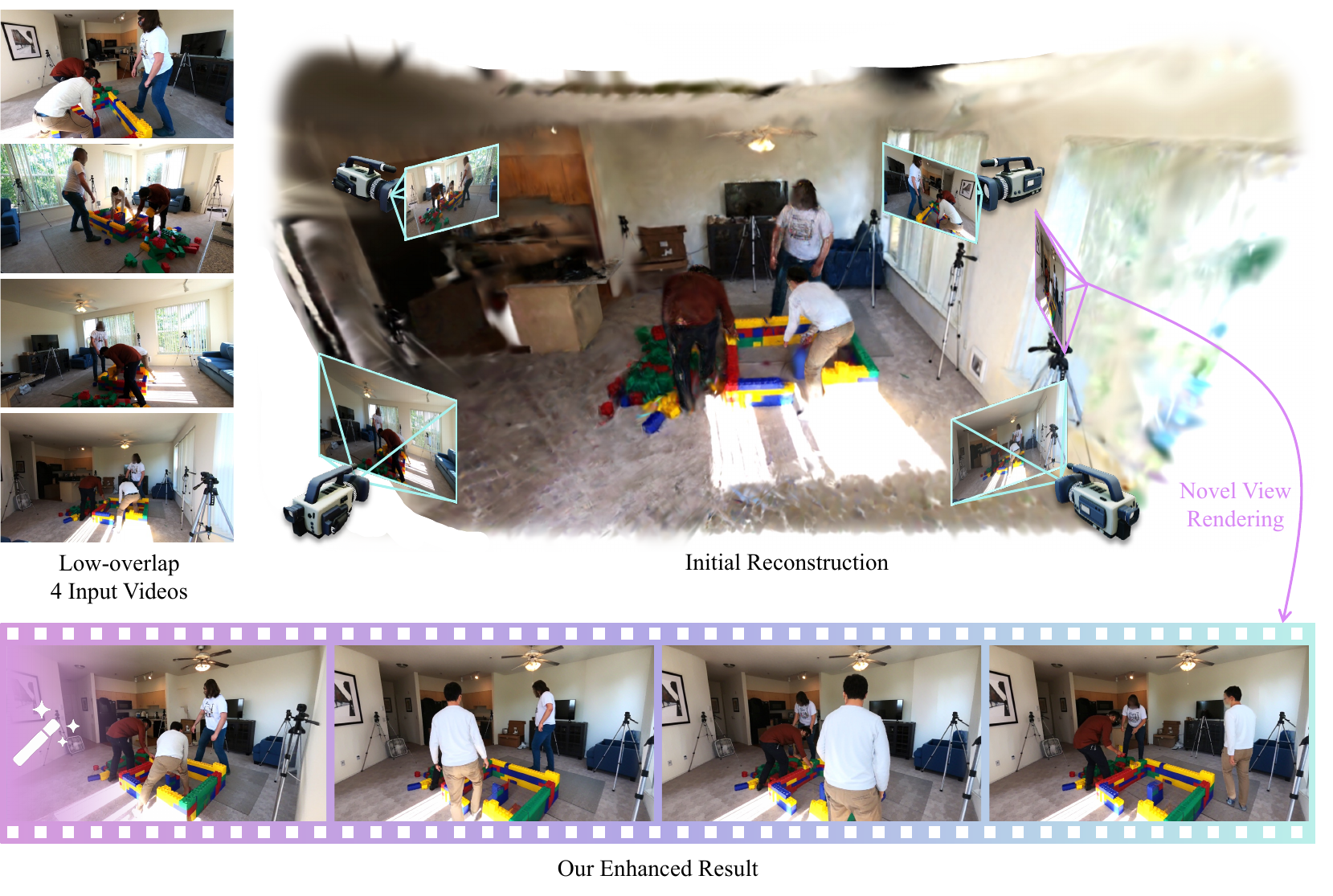}
  \caption{Given only as few as four sparse, low-overlap input videos (left), \ours{} first reconstructs decoupled Gaussians for background and humans (right). The reconstructed Gaussians enable rendering from novel viewpoints, and our recursive enhancement module further refines the rendered output (bottom).}
  \Description{Teaser figure showing input views and novel view synthesis results.}
  \label{fig:teaser}
\end{teaserfigure}

\maketitle

\section{Introduction}

High-fidelity 4D human capture has become essential for entertainment, sports broadcasting, and virtual production.
Professional volumetric systems achieve compelling results but require dozens to hundreds of cameras in controlled environments.
However, in environments such as gymnasiums and homes, only a handful of uncalibrated, low-overlap cameras are available, and these environments often involve multiple interacting people with frequent occlusions.
We refer to these settings as \emph{in-the-wild studio capture}.

Recent advances in 4D Gaussian Splatting have enabled high-quality dynamic scene reconstruction, but predominantly assume dense camera coverage with substantial view overlap~\cite{wu20244d, li2024spacetime, wang2025freetimegs}.
Some sparse-view methods~\cite{li2024dngaussian, park2025dropgaussian, liu2024deceptive, kong2025generative} demonstrate the capability to operate with reduced camera requirements, but still rely on largely shared visibility between neighboring cameras for correspondence matching.
Other works, such as MonoFusion~\cite{wang2025monofusion}, target more challenging low-overlap scenarios (\ie, cameras positioned 90 degrees apart around the scene), yet they still exhibit noticeable artifacts in under-observed regions, as errors in humans and backgrounds become entangled in the shared representation.
Camera-controlled video diffusion models~\cite{bai2025recammaster, ren2025gen3c, jin2025diffuman4d, wu2025cat4d} have emerged as an alternative by directly synthesizing novel views. However, in low-overlap, multi-person settings, they produce geometric inconsistencies for moving people (Appendix~\ref{sec:appendix_results}).

To address these challenges, we present \ours{}, a pipeline that achieves high-fidelity 4D human-scene reconstruction for in-the-wild studio capture.
Our key insight is that backgrounds and humans benefit from different priors.
For background, we must synthesize plausible content beyond the captured views while remaining consistent with the observed evidence, as photometric losses cannot supervise never-observed regions.
Video diffusion models excel at synthesizing plausible static scenes from novel viewpoints by leveraging semantic priors learned from large-scale data.
For humans, however, these models struggle to maintain consistent motion across views, as human motion is inherently complex.
In contrast, parametric body models like SMPL~\cite{loper2023smpl} provide strong geometric priors that constrain human shape and articulation, enabling robust reconstruction even from sparse observations.

This motivates a decoupled reconstruction, avoiding the entanglement causing artifacts in joint methods~\cite{wang2025shape, wang2025monofusion}.
For background reconstruction, we leverage a prior from a camera-controlled video diffusion model~\cite{ren2025gen3c}, synthesizing hundreds of novel views from sparse inputs.
This provides dense supervision for background Gaussians, preventing quality degradation caused by insufficient view coverage.
For human reconstruction, we leverage the body model prior, such as SMPL~\cite{loper2023smpl}.
Our geometry-driven multi-view SMPL estimation associates identities across views via combined spatial and pose affinity, and triangulates multi-view keypoints to recover accurate body parameters even when individual views are occluded.

Compositing separately rendered backgrounds and humans can yield unnatural results with artifacts in under-observed regions, since separate optimization can result in discordant backgrounds and humans.
Therefore, we utilize a single-step diffusion model~\cite{wu2025difix3d+} to remove static Gaussian artifacts, effectively removing them in under-observed regions and improving concordance between backgrounds and humans.
However, directly using a single-step diffusion model can lead to temporal flickering, as this is optimized for per-image refinement.
To address this issue, we build on this single-step diffusion model and propose motion-adaptive consistency injection, which blends each frame with previous outputs warped via optical flow to achieve temporally coherent enhancement.
Experiments on EgoHumans~\cite{khirodkar2023ego}, Harmony4D~\cite{khirodkar2024harmony4d}, Mobile Stage~\cite{xu20244k4d}, and SelfCap~\cite{xu2024representing} demonstrate state-of-the-art novel view synthesis quality.
Our pipeline also enables applications such as novel camera trajectory rendering and human replacement.
To the best of our knowledge, \ours{} is the first pipeline to deliver both high-fidelity human-scene reconstruction and consistent rendering enhancement for in-the-wild studio capture.

In summary, our contributions are fourfold:
\begin{itemize}[leftmargin=1em]
  \item \textbf{Decoupled reconstruction with complementary priors}. Under low-overlap constraints, we observe that backgrounds and humans benefit from different priors. We propose a decoupled approach that uses diffusion models to synthesize dense supervision for backgrounds, while parametric body models constrain human geometry.
  \item \textbf{Geometry-driven multi-view human estimation.} We establish cross-view human correspondences via combined spatial and pose affinity, and triangulate 2D keypoints to obtain 3D body pose, enabling robust human initialization.
  \item \textbf{Motion-adaptive diffusion enhancement}. We apply single-step diffusion with motion-adaptive consistency injection to harmonize separately rendered humans and backgrounds into temporally coherent outputs.
  \item \textbf{State-of-the-art results.} We substantially outperform existing methods across four real-world datasets (EgoHumans, Harmony4D, Mobile Stage, and SelfCap) with diverse camera configurations. We also show applications such as novel camera trajectory rendering and human replacement.
\end{itemize}

\begin{figure*}[t]
  \centering
  \includegraphics[width=0.96\textwidth]{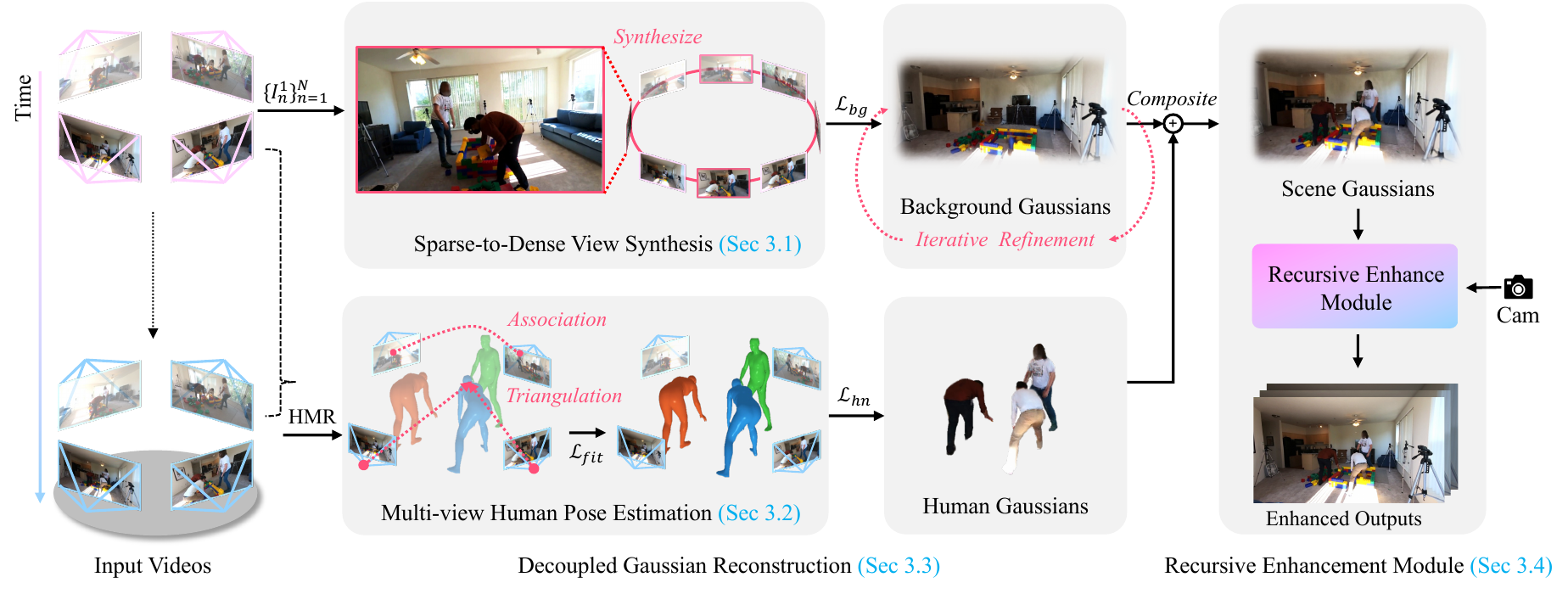}
  \caption{Overview of the proposed \ours{}. Our pipeline consists of four stages: (1) sparse-to-dense view synthesis using camera-controlled video diffusion, (2) multi-view human pose estimation, (3) decoupled Gaussian reconstruction for background and humans, and (4) a recursive enhancement module.}
  \Description{Pipeline overview diagram showing the four stages of our method.}
  \label{fig:pipeline}
\end{figure*}

\section{Related Work}
\subsection{4D Scene Reconstruction}
4D scene reconstruction aims to model the spatio-temporal movements in the scene. Traditional methods rely on dense multi-view setups or RGB-D sensors, as in free-viewpoint video systems~\cite{carranza2003free} and Fusion4D~\cite{dou2016fusion4d}.
With advances in neural rendering such as NeRF~\cite{mildenhall2021nerf}, several works extended them to 4D by modeling space-time dynamics via planar factorization~\cite{fridovich2023k}, multi-view feature aggregation~\cite{li2023dynibar}, and 4D neural voxels~\cite{wu20254d}.
More recently, 3D Gaussian Splatting (3DGS)~\cite{kerbl20233d} has emerged as a faster alternative, where subsequent works extended it to dynamic settings by deforming 3D Gaussians with a deformation field~\cite{luiten2024dynamic, wu20244d, yang2024deformable, huang2024sc} or by representing the scene with 4D Gaussians defined in the spatio-temporal dimension~\cite{li2024spacetime, yang2023gs4d, wang2025freetimegs, duan20244d}.
Despite these advances, most methods assume sufficient view coverage, limiting practical deployment in real-world scenarios.
MonoFusion~\cite{wang2025monofusion} addresses sparse-view inputs, yet artifacts remain in under-observed regions, motivating our decoupled approach with complementary priors for backgrounds and humans.

\subsection{Human-Scene Reconstruction}
Human-scene reconstruction recovers both humans and static scenes from video, enabling applications such as pose editing and novel view synthesis.
Early studies explored disentangled representations, jointly optimizing humans and scenes from monocular video~\cite{kocabas2024hugs,xue2024hsr}.
Subsequent work proposed online methods for simultaneous camera tracking, pose estimation, and reconstruction~\cite{zhang2025odhsr}, extended to multi-view and unconstrained inputs~\cite{kim2025showmak3r,huang2025echoes, moon2024expressive}.
Zhan~\textit{et al.}~\cite{zhan2025towards} model humans holding objects via compositional representations, though they require about 24 views. Diffusion-based approaches~\cite{jin2025diffuman4d, shao2024360} generate multi-view consistent humans but focus on individual humans without background scenes.

\subsection{Diffusion-Based Video Enhancement}
Diffusion models provide strong priors for video enhancement, but temporal coherence remains challenging due to stochastic denoising, which causes flicker and texture drift. Prior work improves consistency by warping diffusion noise across time~\cite{chang2025warped,daras2024warped}.
Task-specific methods incorporate temporal modules~\cite{zhou2024upscale,yang2024motion}, while training-free approaches adapt image diffusion models via latent warping~\cite{yeh2024diffir2vr}.
Diffusion priors have also been used to enhance neural rendering outputs.
3DGS-Enhancer~\cite{liu20243dgs} applies diffusion to enhance 3DGS, and Difix3D+~\cite{wu2025difix3d+} explores single-step diffusion for artifact removal.

However, two gaps remain: (i) in-the-wild studio capture suffers from missing view coverage before refinement, and (ii) free-viewpoint rendering requires efficient, temporally-stable enhancement where multi-step diffusion is costly and per-frame enhancement causes flicker.
We address these by synthesizing dense novel views via video diffusion for background supervision, and applying single-step diffusion with motion-adaptive consistency injection for artifact-free rendering.

\section{Method}
\label{sec:method}

We study the problem of reconstructing dynamic scenes with multiple humans from $N$ synchronized cameras over $T$ timesteps.
What we aim to solve is \emph{in-the-wild studio capture}, characterized by sparse camera layouts with low-overlap scenarios (\ie, neighboring cameras observe largely disjoint regions), multiple interacting subjects, and occlusions.
Such conditions commonly arise in sports venues, healthcare facilities, and home environments.
Additionally, we do not assume pre-calibrated cameras; instead, camera parameters are estimated using a feed-forward reconstruction model (Section~\ref{sec:view_synthesis}).

Given input videos $\{I_n^t\}_{n=1,t=1}^{N,T}$ where $I_n^t \in \mathbb{R}^{H \times W \times 3}$, our goal is two-fold:
(1) obtain disentangled 3DGS-based representations of static background and dynamic humans, and
(2) produce consistent video renderings from arbitrary camera trajectories.
This setting introduces two key challenges: sparse viewpoint coverage leaves large portions of the scene unobserved, and reconstructing dynamic humans is ill-posed under limited observations with self-occlusions.
Our insight is that these challenges are best addressed with distinct priors.
Video diffusion models can synthesize dense viewpoints for backgrounds, while parametric body models (e.g., SMPL) impose geometric constraints on humans.
This motivates a \emph{decoupled reconstruction} strategy.

As illustrated in Figure~\ref{fig:pipeline}, our pipeline consists of four stages:
(1) sparse-to-dense view synthesis to compensate for insufficient viewpoint coverage from sparse cameras;
(2) multi-view human pose estimation to establish geometric priors that constrain the ill-posed human reconstruction;
(3) decoupled Gaussian reconstruction, optimizing backgrounds on synthesized views and humans on original videos, where each supervision source is most reliable; and
(4) recursive enhancement module to harmonize separately optimized components and resolve remaining artifacts.
\subsection{Sparse-to-Dense View Synthesis}
\label{sec:view_synthesis}

Compared to the conventional reconstruction setting, sparse cameras provide insufficient viewpoint coverage for 3DGS.
We synthesize dense novel views to augment the limited observations.

\paragraph{Preprocessing.}
Since cameras are static, we apply a feed-forward 3D reconstruction model~\cite{wang2025pi} to the first frame ($t=1$) from all $N$ cameras to obtain dense point clouds, depth maps, and camera poses that are reused for all timesteps.
For subsequent timesteps, we estimate depth using a monocular model~\cite{wang2025moge} aligned to the first-frame depth for scale consistency.
To initialize background Gaussians without human contamination, we segment humans using a segmentation model~\cite{carion2025sam} to obtain masks $\{M_n^1\}_{n=1}^{N}$, and extract the background point cloud by retaining only unmasked pixels with depth confidence above a threshold.

\paragraph{Camera Trajectory Generation.}
We interpolate between the $N$ input viewpoints using spherical linear interpolation (SLERP)~\cite{jang2024spherical} for rotation and linear interpolation for translation, yielding $L$ novel camera poses (Appendix~\ref{sec:appendix_preprocessing}).

\paragraph{Novel View Synthesis.}
We use a video diffusion model~\cite{ren2025gen3c} to synthesize novel views along the interpolated camera trajectory, taking $N$ first-frame images, depth maps, and target poses as input.
We also obtain human masks for each synthesized view, which are used to exclude human regions from these images during background Gaussian optimization (Section~\ref{sec:gaussian}).


\begin{figure}[t]
\centering
\includegraphics[width=0.9\linewidth]{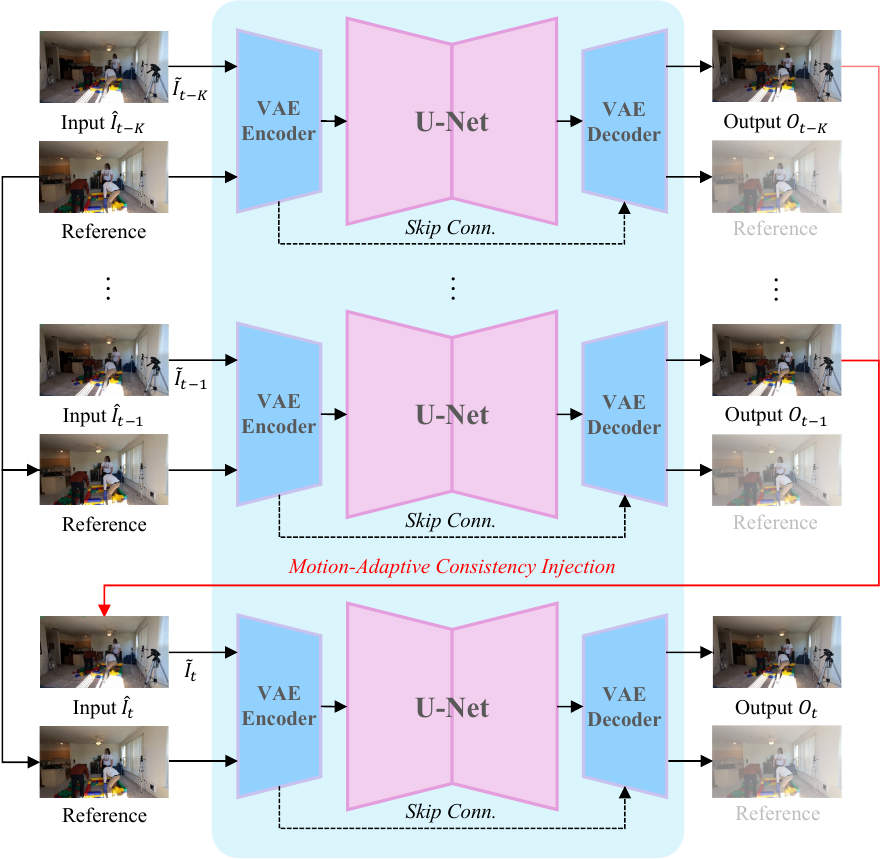}
\caption{Overview of our recursive enhancement module (Sec.~\ref{sec:refinement}).}
\Description{Diagram showing the recursive enhancement module pipeline: rendered composite and reference frame are input to a single-step diffusion model, with motion-adaptive consistency injection blending warped previous outputs via optical flow for temporal coherence.}
\label{fig:enhance_module}
\end{figure}

\subsection{Multi-View Human Pose Estimation}
\label{sec:human_init}

Accurate human reconstruction requires estimating body pose in a unified world coordinate system, which requires establishing correspondence between detections of the same person across views.
In in-the-wild studio capture, conventional appearance-based matching becomes unreliable as appearance varies significantly across viewpoints.
We address this by detecting and tracking humans independently in each view, associating identities across views using geometric cues, and triangulating multi-view observations to estimate 3D body parameters.

\subsubsection{Per-View Human Detection}

For each view $n$, we apply a monocular pose estimator~\cite{newell2025comotion} that provides SMPL pose $\boldsymbol{\theta}$ and shape $\boldsymbol{\beta}$ estimates, 2D keypoints, and within-view identity tracking across all timesteps.
The monocular SMPL estimates are used to guide cross-view identity association, but not for final reconstruction, as direct 3D regression suffers from depth ambiguity.

\subsubsection{Cross-View Identity Association}

A central challenge of multi-view multi-human pose estimation is cross-view identity association. Unlike dense setups where visual overlap enables appearance matching, our cameras observe largely disjoint regions.
We leverage calibrated camera geometry to lift detections to 3D and match based on spatial proximity and pose similarity.

For each detected person, we estimate their 3D world position $\mathbf{p}$ by unprojecting the 2D pelvis location $(u, v)$ using the depth and camera parameters from Section~\ref{sec:view_synthesis}.

We define the affinity between two detections $a$ and $b$ from different views as a weighted combination of spatial and pose similarities:
\begin{equation}
A(a, b) = w_p \cdot \exp\left(-\frac{\|\mathbf{p}^a - \mathbf{p}^b\|}{\sigma_p}\right) + w_\theta \cdot \exp\left(-\frac{\|\boldsymbol{\theta}^a - \boldsymbol{\theta}^b\|}{\sigma_\theta}\right)
\end{equation}
where $\mathbf{p}$ denotes the 3D world position, $\boldsymbol{\theta}$ the SMPL pose parameters, $\sigma$ the scale parameters, and we set $w_p = 0.9$, $w_\theta = 0.1$.
This combination provides robustness when depth is noisy or when people perform similar actions.

Starting from the reference timestep $t_0$, we apply the Hungarian algorithm to assign cross-view correspondences based on the affinity measure, while applying a distance threshold to reject background false positives.
This produces a mapping from global person identities to view-specific detection indices. For temporal consistency, we primarily rely on the within-view tracking provided by the monocular estimator. When a person reappears after occlusion with a new track ID, we reassign the global identity using the affinity measure.

\subsubsection{3D Pose Triangulation}
\label{sec:triangulation}

Given cross-view correspondences, we triangulate the 2D keypoints across all views to obtain 3D joint and vertex positions for SMPL fitting.
We use robust optimization with Huber loss to handle occlusions and detection noise.
For each point $k$, we solve for the 3D world position that minimizes reprojection error across all $N$ views:
\begin{equation}
\mathbf{P}^{w}_k = \argmin_{\mathbf{X}} \sum_{n=1}^{N} \rho\left(\|\mathbf{x}_{n,k} - \pi_n(\mathbf{X})\|\right)
\end{equation}
where $\mathbf{P}^{w}_k$ denotes the triangulated 3D position, $\mathbf{x}_{n,k}$ is the corresponding 2D keypoint (joint or vertex) in view $n$, $\pi_n$ is the projection function, and $\rho$ is the Huber loss for robustness to outliers.

The triangulated 3D points are in the scene coordinate system. We compute a scale factor by matching SMPL skeleton template bone lengths against the triangulated bone lengths:
\begin{equation}
s^* = \argmin_s \sum_{(i,j) \in \mathcal{B}} \rho\left(\|\mathbf{J}^{w}_i - \mathbf{J}^{w}_j\| - s \cdot \|\mathbf{J}^{s}_i - \mathbf{J}^{s}_j\|\right)
\end{equation}
where $(i,j)$ are adjacent joint indices forming a bone, $\mathcal{B}$ is the set of reliable bone pairs, $\mathbf{J}^{w}$ are the triangulated world-space joints, and $\mathbf{J}^{s}$ are template SMPL joint positions.
We compute a global scale per person by aggregating across frames.

\subsubsection{SMPL Parameter Fitting}

We fit SMPL parameters to the triangulated 3D keypoints (Section~\ref{sec:triangulation}) using 3D-to-3D optimization~\cite{javerliat2025kineo}, which avoids depth ambiguities inherent in 2D projection fitting.
We optimize pose $\boldsymbol{\theta}$ and shape $\boldsymbol{\beta}$ to minimize:
\begin{equation}
\begin{split}
\mathcal{L}_{fit} = &\sum_{j} \|s^* \mathbf{J}^{s}_j(\boldsymbol{\theta}, \boldsymbol{\beta}) - \mathbf{J}^{w}_j\|^2 \\
&+ \sum_{i} \|s^* \mathbf{V}^{s}_i(\boldsymbol{\theta}, \boldsymbol{\beta}) - \mathbf{V}^{w}_i\|^2 + \lambda_\beta \|\boldsymbol{\beta}\|^2
\end{split}
\end{equation}
where $\mathbf{J}^{s}(\boldsymbol{\theta}, \boldsymbol{\beta})$ and $\mathbf{V}^{s}(\boldsymbol{\theta}, \boldsymbol{\beta})$ are the joints and vertices produced by the SMPL body model, and $\mathbf{J}^{w}$ and $\mathbf{V}^{w}$ are the triangulated observations.
The joint term aligns the skeleton pose, while the vertex term provides surface-level supervision, enabling body shape estimation. Finally, we apply exponential smoothing to the fitted pose parameters for temporal consistency.

\subsection{Decoupled Gaussian Reconstruction}
\label{sec:gaussian}

Having obtained camera parameters, dense novel views, and SMPL parameters, we reconstruct the scene as the composition of a static background and dynamic human Gaussians.
Backgrounds and humans benefit from different optimization strategies: backgrounds require dense viewpoint coverage but only a single timestep, while humans require temporal supervision but can be constrained by SMPL priors despite partial observation.
We decouple and reconstruct backgrounds and humans separately.

\subsubsection{Background Reconstruction}

Using the point cloud, $L$ synthesized novel views, and corresponding human masks obtained in Section~\ref{sec:view_synthesis}, we initialize and optimize background Gaussians at timestep $t=1$.
Since synthesized views contain humans, we mask these regions during optimization:
\begin{equation}
\mathcal{L}_{bg} = (1 - M) \odot \left(\mathcal{L}_1 + \lambda_s \mathcal{L}_{SSIM} + \lambda_l \mathcal{L}_{LPIPS}\right) + \lambda_d \mathcal{L}_{den}
\end{equation}
where $M$ is the dilated human mask and $\mathcal{L}_{den}$ regularizes opacity to prevent floaters.
We apply higher loss weights to views near original camera positions using cosine falloff based on angular distance, as these provide ground-truth supervision.

\paragraph{Iterative Refinement.}
The synthesized views may undersample certain regions, particularly floors and ceilings at oblique angles.
At the midpoint of optimization, we create additional supervision by rendering the current Gaussians from novel viewpoints with height variation, adding sinusoidal vertical displacement, and compensating pitch rotation. Human masks are rendered from SMPL meshes.
We refine these rendered images using the recursive enhancement module (Section~\ref{sec:refinement}), which applies single-step diffusion to correct artifacts. These refined images serve as additional supervision, improving quality in undersampled regions (Figure~\ref{fig:iterative_qual}).
We provide runtime analysis in Appendix~\ref{sec:appendix_runtime}. After background optimization, the Gaussians are frozen for human reconstruction.

\begin{figure}[tp]
\centering
\includegraphics[width=0.83\linewidth]{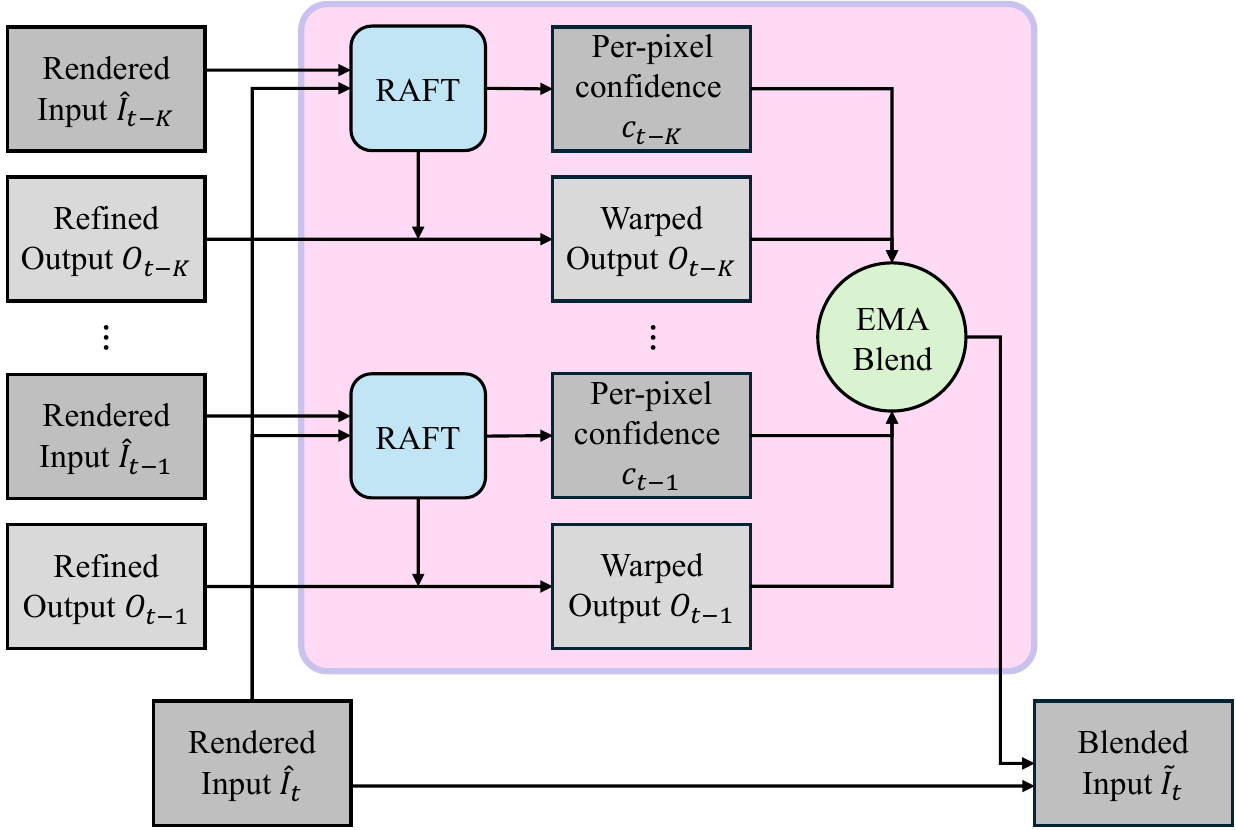}
\caption{Schematic of motion-adaptive consistency injection. For each frame, RAFT computes backward flow to warp previous enhanced outputs, which are blended with the current input via per-pixel confidence-weighted EMA before single-step diffusion.}
\Description{Schematic diagram of EMA consistency injection.}
\label{fig:ema_schematic_main}
\end{figure}

\subsubsection{Human Reconstruction}

Initialized with SMPL parameters from Section~\ref{sec:human_init}, we optimize human Gaussians on the original $N$ input videos across all $T$ timesteps, jointly refining pose and learning dynamic appearance from temporal variation.

\paragraph{Canonical Human Representation.}
Following recent work~\cite{hu2024gauhuman, kocabas2024hugs}, each person is represented by Gaussians in a canonical pose, initialized on the SMPL mesh surface~\cite{loper2023smpl}.
The canonical Gaussians capture person-specific appearance, while pose-dependent deformation is handled by skeletal skinning.

\paragraph{Skeletal Deformation.}
To render a person at timestep $t$, we deform the canonical Gaussians to the target pose using Linear Blend Skinning (LBS)~\cite{loper2023smpl}.
For each Gaussian center $\boldsymbol{\mu}_c$ in canonical space, the deformed position is computed as:
\begin{equation}
\boldsymbol{\mu}_t = \sum_{b=1}^{B} w_b(\boldsymbol{\mu}_c) \cdot \mathbf{G}_b^t \cdot \boldsymbol{\mu}_c
\end{equation}
where $w_b(\boldsymbol{\mu}_c)$ are skinning weights queried from a precomputed voxel grid derived from the SMPL model, and $\mathbf{G}_b^t \in SE(3)$ are the bone transformation matrices for timestep $t$.
To capture fine-grained appearance changes, we employ a temporal MLP that predicts per-Gaussian color and opacity residuals.
We jointly optimize Gaussian attributes and SMPL pose parameters using $\mathcal{L}_{hn}$, which combines photometric losses (L1, SSIM, LPIPS), silhouette supervision, density regularization to prevent spurious Gaussians, and temporal smoothness constraints on SMPL parameters.
Details of the objective are provided in Appendix~\ref{sec:appendix_human}.

\begin{figure*}[t]
\centering
\includegraphics[width=1\textwidth]{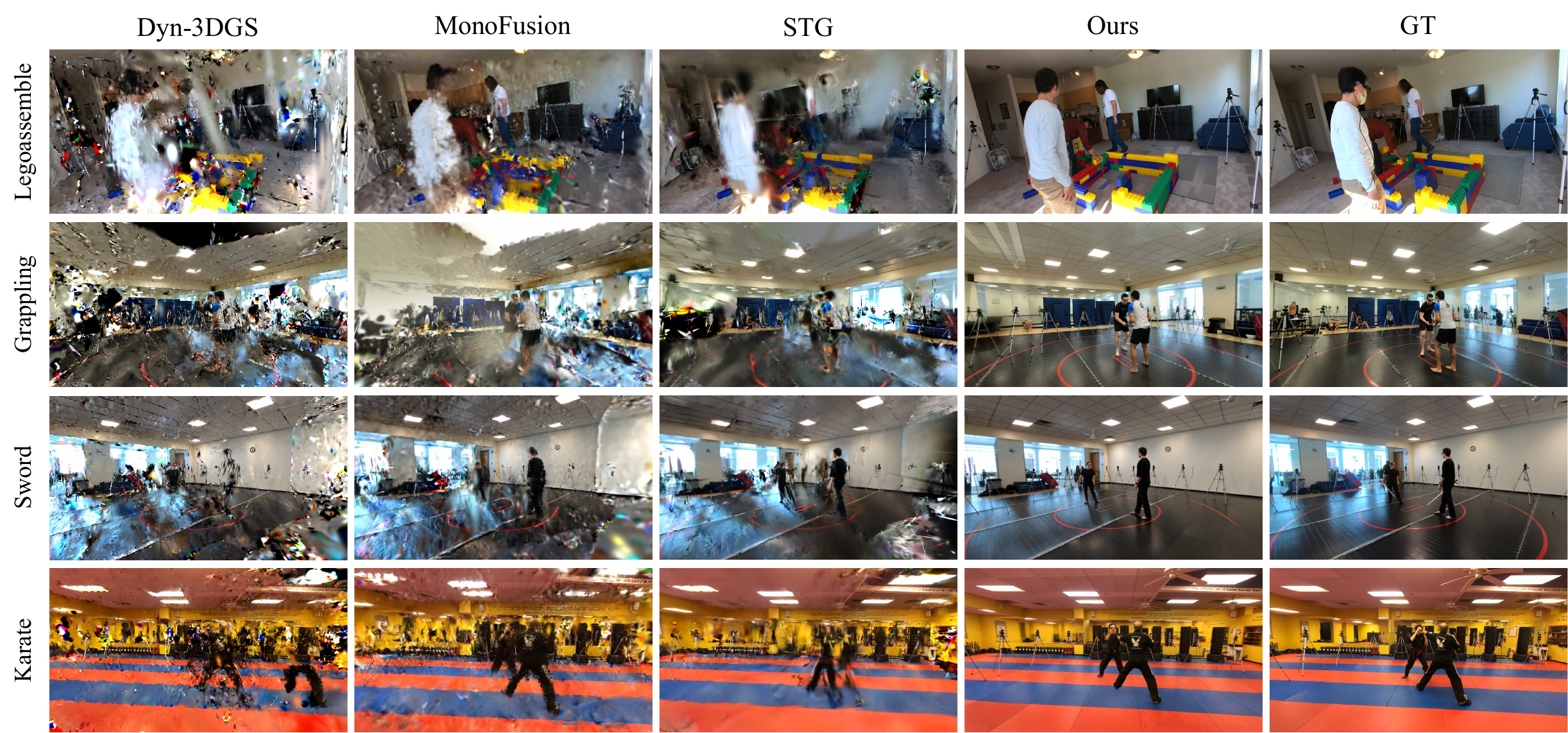}
\caption{Qualitative comparison on 360\degree{} scenes (\textit{Legoassemble}, \textit{Grappling}, \textit{Sword}, \textit{Karate}). Our method produces sharper backgrounds and more robust human reconstructions than baselines.}
\Description{Qualitative comparison on 360-degree scenes.}
\label{fig:qual_main}
\end{figure*}

\subsection{Recursive Enhancement Module}
\label{sec:refinement}

Despite achieving superior reconstruction quality compared to existing methods~\cite{luiten2024dynamic, wang2025monofusion, li2024spacetime}, rendered outputs from the reconstructed Gaussians may still exhibit minor artifacts such as floaters, blurring in under-observed regions, and human artifacts from pose inaccuracies.
Since humans and backgrounds are optimized separately, texture and lighting may also vary slightly when composited.

We address these using a single-step diffusion model designed for Gaussian refinement~\cite{wu2025difix3d+}, which takes two inputs: the rendered image and a reference image providing appearance guidance.
For each frame, we use the rendered composite as input and a nearby raw frame as reference (Figure~\ref{fig:enhance_module}).

\paragraph{Motion-Adaptive Consistency Injection.}
Applying the model independently to each frame causes flickering. Inspired by shared noise priors for video generation~\cite{wang2024microcinema}, we observe that since the single-step model treats the input as the noise-corrupted sample at timestep $t=200$, injecting information from previous frames is equivalent to injecting a shared noise prior. As validated in our ablation study, we blend each rendered input with warped previous outputs using an exponential moving average (EMA) with motion-adaptive weighting:
\begin{equation}
\tilde{I}_t = (1 - w_{total}) \cdot \hat{I}_t + w_{total} \cdot \sum_{i=0}^{K-1} \bar{w}_i \cdot c_i \cdot \text{warp}(O_{t-1-i}, F_{t \rightarrow t-1-i})
\end{equation}
where $\tilde{I}_t$ is the blended input fed to the diffusion model, $\hat{I}_t$ is the rendered composite, and $w_{total}$ is the overall injection strength.
$O_{t-1-i}$ denotes the refined output from $i+1$ frames ago, with $\bar{w}_i$ being normalized EMA weights that apply exponential decay over $K$ previous frames.
$F_{t \rightarrow t-1-i}$ is the backward optical flow computed using RAFT~\cite{teed2020raft}, and $c_i = \max(0, 1 - e_i/\tau_e)$ is per-pixel confidence based on warp error $e_i$ for each previous frame.
The exponential decay prioritizes recent frames, which have better flow accuracy, while incorporating multiple frames provides robustness.
Static regions where the warp aligns well receive full injection ($c_i \approx 1$), while moving regions where the warp fails receive minimal injection ($c_i \approx 0$) to prevent ghosting.
Figure~\ref{fig:ema_schematic_main} illustrates the overall mechanism, and Figure~\ref{fig:flow_diagram} visualizes intermediate results on an example frame.

\section{Experiment}
\subsection{Setup}

We evaluate on two camera configurations: 360{\degree} coverage using EgoHumans~\cite{khirodkar2023ego} and Harmony4D~\cite{khirodkar2024harmony4d}, and 180{\degree} coverage using Mobile Stage~\cite{xu20244k4d} and SelfCap~\cite{xu2024representing} where cameras cover only the frontal hemisphere.
For all datasets, we select 4 training cameras spaced approximately 90{\degree} apart and evaluate on 4 other cameras at intermediate angles (approximately 45{\degree}).
Each sequence contains 121 frames with 1-3 people performing dynamic activities.
We compare against Dyn-3DGS~\cite{luiten2024dynamic}, MonoFusion~\cite{wang2025monofusion}, and STG~\cite{li2024spacetime}, all trained on the 4 sparse input views.
Since COLMAP~\cite{schonberger2016structure} fails under sparse settings, we replace it with a feed-forward model~\cite{wang2025pi} for baseline initialization.
We report PSNR, SSIM~\cite{nilsson2020understanding}, and LPIPS~\cite{zhang2018unreasonable} on held-out cameras. Implementation details are provided in Appendices~\ref{sec:appendix_preprocessing}--\ref{sec:appendix_refinement}. Appendix~\ref{sec:appendix_ablation} provides robustness analysis on multi-view pose refinement, temporal smoothing, and mask dilation.

\begin{table*}[t]
\centering
\caption{Quantitative comparison on 360{\degree} scenes. \colorbox{red!20}{red}: best, \colorbox{yellow!50}{yellow}: second best.}
\label{tab:360}
\setlength{\tabcolsep}{2.0pt}
\resizebox{\textwidth}{!}{
\begin{tabular}{l|ccc|ccc|ccc|ccc|ccc|ccc}
\toprule
& \multicolumn{9}{c|}{EgoHumans} & \multicolumn{9}{c}{Harmony4D} \\
& \multicolumn{3}{c|}{Legoassemble} & \multicolumn{3}{c|}{Tennis} & \multicolumn{3}{c|}{Fencing} & \multicolumn{3}{c|}{Sword} & \multicolumn{3}{c|}{Karate} & \multicolumn{3}{c}{Grappling} \\
& PSNR$\uparrow$ & SSIM$\uparrow$ & LPIPS$\downarrow$ & PSNR$\uparrow$ & SSIM$\uparrow$ & LPIPS$\downarrow$ & PSNR$\uparrow$ & SSIM$\uparrow$ & LPIPS$\downarrow$ & PSNR$\uparrow$ & SSIM$\uparrow$ & LPIPS$\downarrow$ & PSNR$\uparrow$ & SSIM$\uparrow$ & LPIPS$\downarrow$ & PSNR$\uparrow$ & SSIM$\uparrow$ & LPIPS$\downarrow$ \\
\midrule
Dyn-3DGS~\cite{luiten2024dynamic} & 15.32 & 0.329 & 0.594 & 17.03 & 0.477 & 0.670 & 18.00 & 0.651 & 0.586 & 16.75 & 0.422 & 0.514 & 16.69 & 0.465 & 0.510 & 16.15 & 0.430 & 0.508 \\
MonoFusion~\cite{wang2025monofusion} & \cellcolor{yellow!50}16.12 & 0.406 & 0.624 & 17.08 & \cellcolor{yellow!50}0.526 & 0.733 & \cellcolor{yellow!50}20.09 & 0.598 & \cellcolor{yellow!50}0.472 & \cellcolor{yellow!50}17.01 & 0.405 & 0.506 & 15.44 & 0.400 & 0.598 & 16.48 & 0.410 & 0.514 \\
STG~\cite{li2024spacetime} & 15.88 & \cellcolor{yellow!50}0.416 & \cellcolor{yellow!50}0.551 & \cellcolor{yellow!50}17.80 & 0.511 & \cellcolor{yellow!50}0.505 & 18.84 & \cellcolor{yellow!50}0.692 & 0.517 & 16.44 & \cellcolor{yellow!50}0.514 & \cellcolor{yellow!50}0.503 & \cellcolor{yellow!50}16.94 & \cellcolor{yellow!50}0.579 & \cellcolor{yellow!50}0.485 & \cellcolor{yellow!50}16.55 & \cellcolor{yellow!50}0.516 & \cellcolor{yellow!50}0.399 \\
Ours & \cellcolor{red!20}18.58 & \cellcolor{red!20}0.569 & \cellcolor{red!20}0.251 & \cellcolor{red!20}19.27 & \cellcolor{red!20}0.578 & \cellcolor{red!20}0.339 & \cellcolor{red!20}22.75 & \cellcolor{red!20}0.748 & \cellcolor{red!20}0.164 & \cellcolor{red!20}20.14 & \cellcolor{red!20}0.648 & \cellcolor{red!20}0.206 & \cellcolor{red!20}19.90 & \cellcolor{red!20}0.688 & \cellcolor{red!20}0.160 & \cellcolor{red!20}19.50 & \cellcolor{red!20}0.646 & \cellcolor{red!20}0.204 \\
\bottomrule
\end{tabular}
}
\end{table*}

\begin{table}[t]
\centering
\caption{Quantitative comparison on 180{\degree} scenes. \colorbox{red!20}{red}: best, \colorbox{yellow!50}{yellow}: second best.}
\label{tab:180}
\resizebox{\linewidth}{!}{
\begin{tabular}{l|ccc|ccc}
\toprule
& \multicolumn{3}{c|}{Mobile Stage} & \multicolumn{3}{c}{SelfCap} \\
& \multicolumn{3}{c|}{Dance} & \multicolumn{3}{c}{Yoga} \\
& PSNR$\uparrow$ & SSIM$\uparrow$ & LPIPS$\downarrow$ & PSNR$\uparrow$ & SSIM$\uparrow$ & LPIPS$\downarrow$ \\
\midrule
Dyn-3DGS~\cite{luiten2024dynamic} & 16.63 & 0.289 & 0.601 & 16.89 & 0.461 & 0.521 \\
MonoFusion~\cite{wang2025monofusion} & \cellcolor{yellow!50}16.73 & 0.298 & \cellcolor{yellow!50}0.516 & 17.49 & 0.436 & 0.452 \\
STG~\cite{li2024spacetime} & 16.52 & \cellcolor{yellow!50}0.336 & 0.563 & \cellcolor{yellow!50}18.72 & \cellcolor{yellow!50}0.609 & \cellcolor{yellow!50}0.289 \\
Ours & \cellcolor{red!20}21.74 & \cellcolor{red!20}0.575 & \cellcolor{red!20}0.145 & \cellcolor{red!20}21.63 & \cellcolor{red!20}0.740 & \cellcolor{red!20}0.115 \\
\bottomrule
\end{tabular}
}
\end{table}

\subsection{Comparison}
Tables~\ref{tab:360} and~\ref{tab:180} compare our method against baselines on 360{\degree} and 180{\degree} camera configurations, respectively.
Our method consistently outperforms all baselines across all metrics, with particularly large LPIPS improvements indicating better perceptual quality.
The largest improvements occur when cameras are placed relatively far apart, leaving portions of the scene poorly captured.
Baselines suffer from insufficient viewpoint coverage, while our dense view synthesis addresses this limitation.

\begin{figure*}[p]
\centering
\includegraphics[width=0.95\textwidth]{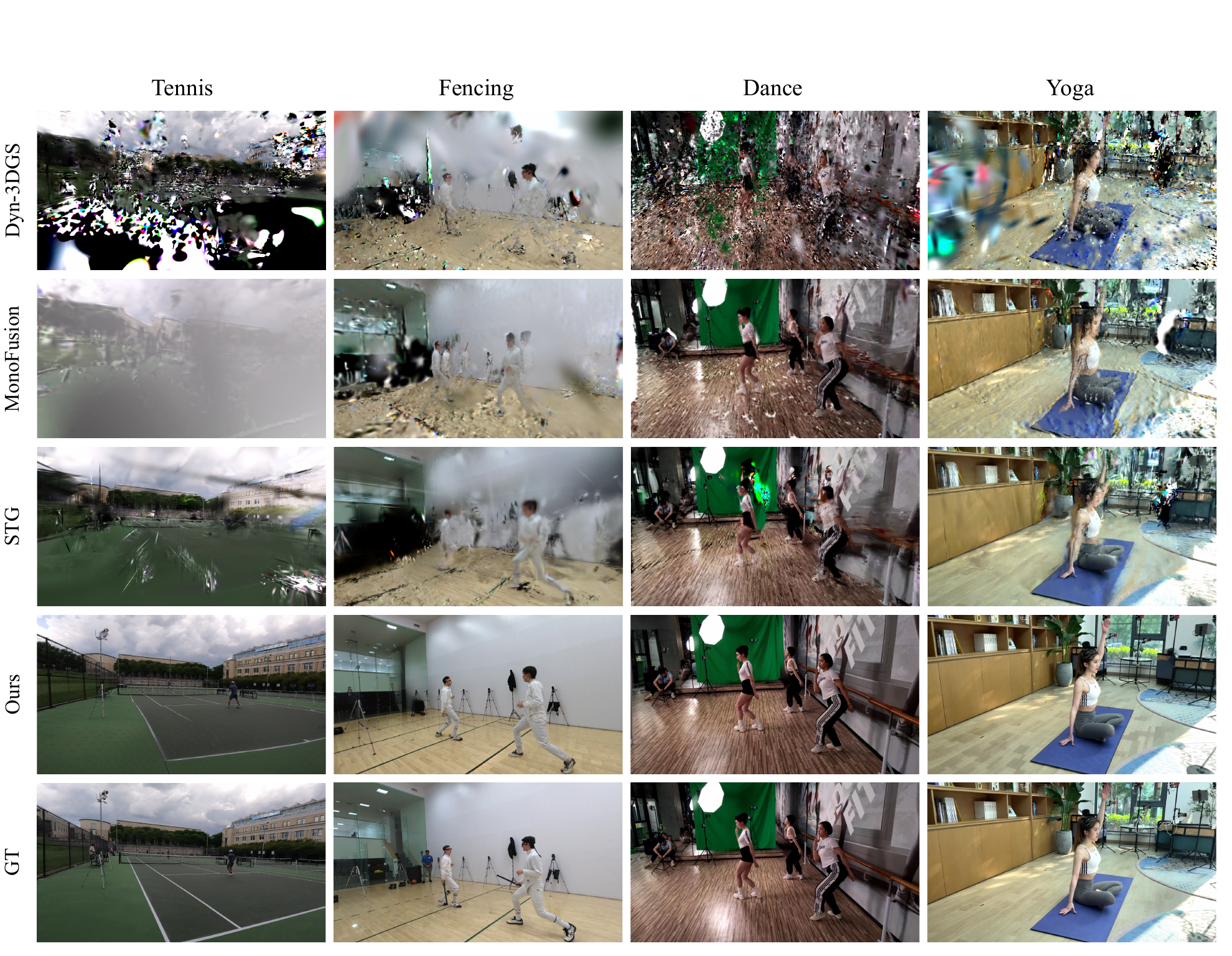}
\caption{Additional qualitative comparison (\textit{Tennis}, \textit{Fencing}, \textit{Dance}, \textit{Yoga}) from EgoHumans, Mobile Stage, and SelfCap. Our method produces sharper reconstructions with better human-scene separation. \textit{Dance} \textcopyright\ Xu et al.\ (Mobile Stage); \textit{Yoga} \textcopyright\ Xu et al.\ (SelfCap), used with permission.}
\Description{Extended qualitative comparison across multiple datasets.}
\label{fig:qual_extended}
\end{figure*}

Qualitative comparisons are shown in Figure~\ref{fig:qual_main} (additional scenes in Figure~\ref{fig:qual_extended}).
Baseline methods exhibit noticeable artifacts in background regions due to insufficient viewpoint coverage and unstable human geometry from limited multi-view constraints.
In contrast, our method produces clean backgrounds and robust human reconstructions by leveraging dense synthesized views for scene modeling and explicit SMPL-guided human representation. See our project page for video results.

\subsection{Ablation Study}

\paragraph{Cross-View Association.}
Table~\ref{tab:ablation_association} validates our hybrid cross-view identity association strategy. We evaluate across 8 scenes by comparing predicted associations against manual ground-truth annotations.
Using only spatial proximity fails when depth is unreliable (93.3\%), while only pose similarity fails when humans perform similar actions (81.4\%). Our hybrid approach combines both cues, achieving 97.8\% accuracy with zero false positives.

\begin{table}[t]
\centering
\caption{Ablation on cross-view actor association strategy averaged across all 8 scenes (Tables~\ref{tab:360},~\ref{tab:180}).}
\label{tab:ablation_association}
\begin{tabular}{l|cc}
\toprule
& Accuracy$\uparrow$ & Precision$\uparrow$ \\
\midrule
Only spatial & 93.3\% & \textbf{100\%} \\
Only pose & 81.4\% & 99.7\% \\
Ours & \textbf{97.8\%} & \textbf{100\%} \\
\bottomrule
\end{tabular}
\end{table}

\begin{table}[t]
\centering
\caption{Ablation on pipeline components averaged across all 8 scenes (Tables~\ref{tab:360},~\ref{tab:180}).}
\label{tab:ablation_components}
\begin{tabular}{l|ccc}
\toprule
& PSNR$\uparrow$ & SSIM$\uparrow$ & LPIPS$\downarrow$ \\
\midrule
4-view baseline & 18.13 & 0.567 & 0.429 \\
+ ViewSynth & 20.54 & 0.661 & 0.273 \\
+ Enhance (ours) & 20.44 & 0.649 & \textbf{0.198} \\
\bottomrule
\end{tabular}
\end{table}

\begin{table}[t]
\centering
\caption{Ablation on motion-adaptive consistency injection averaged across all 8 scenes (Tables~\ref{tab:360},~\ref{tab:180}). Warp-L2 measures the L2 error between consecutive frames after optical flow warping.}
\label{tab:ablation_inject}
\begin{tabular}{l|cccc}
\toprule
& PSNR$\uparrow$ & SSIM$\uparrow$ & LPIPS$\downarrow$ & Warp-L2$\downarrow$ \\
\midrule
w/o Injection & 20.44 & 0.654 & \textbf{0.194} & 0.119 \\
w/ Injection & 20.44 & 0.649 & 0.198 & \textbf{0.092} \\
\bottomrule
\end{tabular}
\end{table}

\begin{figure*}[p]
  \centering
  \includegraphics[width=0.95\linewidth]{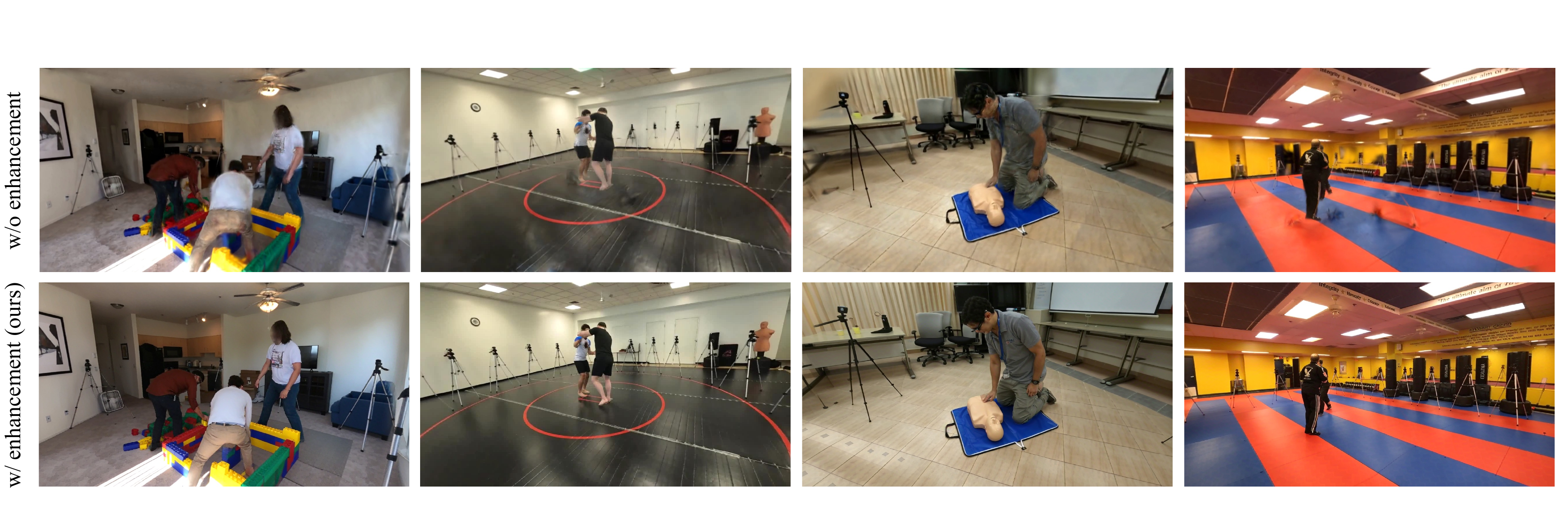}
  \caption{Ablation on recursive enhancement (Sec.~\ref{sec:refinement}). Raw renders (left) contain blur and geometric instabilities. Enhancement (right) produces clean, harmonized outputs.}
  \Description{Ablation study comparing raw renders with refined outputs.}
  \label{fig:ablation_refine}
\end{figure*}

\paragraph{Pipeline Components.}
Table~\ref{tab:ablation_components} ablates our two key contributions: dense view synthesis via video diffusion and single-step diffusion enhancement.
Dense view synthesis provides the largest gain (+2.4 PSNR, 36\% LPIPS reduction), demonstrating that video diffusion effectively addresses sparse-view challenges. The enhancement further improves perceptual quality (27\% LPIPS reduction), while the minor PSNR/SSIM drop reflects generative details that are not pixel-aligned with the ground truth.
As shown in Figure~\ref{fig:ablation_refine}, raw Gaussian outputs suffer from blur and instabilities in under-observed regions, and decoupled reconstruction can cause humans and backgrounds to appear disjointed. The enhancement resolves both issues, producing clean, harmonized outputs.

\paragraph{Motion-Adaptive Consistency Injection.}
When diffusion enhancement is applied independently per frame, ambiguous regions are resolved differently across frames, causing flickering. Our motion-adaptive consistency injection blends warped previous outputs into the current input, encouraging consistent ambiguity resolution across time. Table~\ref{tab:ablation_inject} evaluates this design.

\begin{figure*}[p]
  \centering
  \begin{minipage}[b]{0.49\textwidth}
    \centering
    \includegraphics[width=1\linewidth]{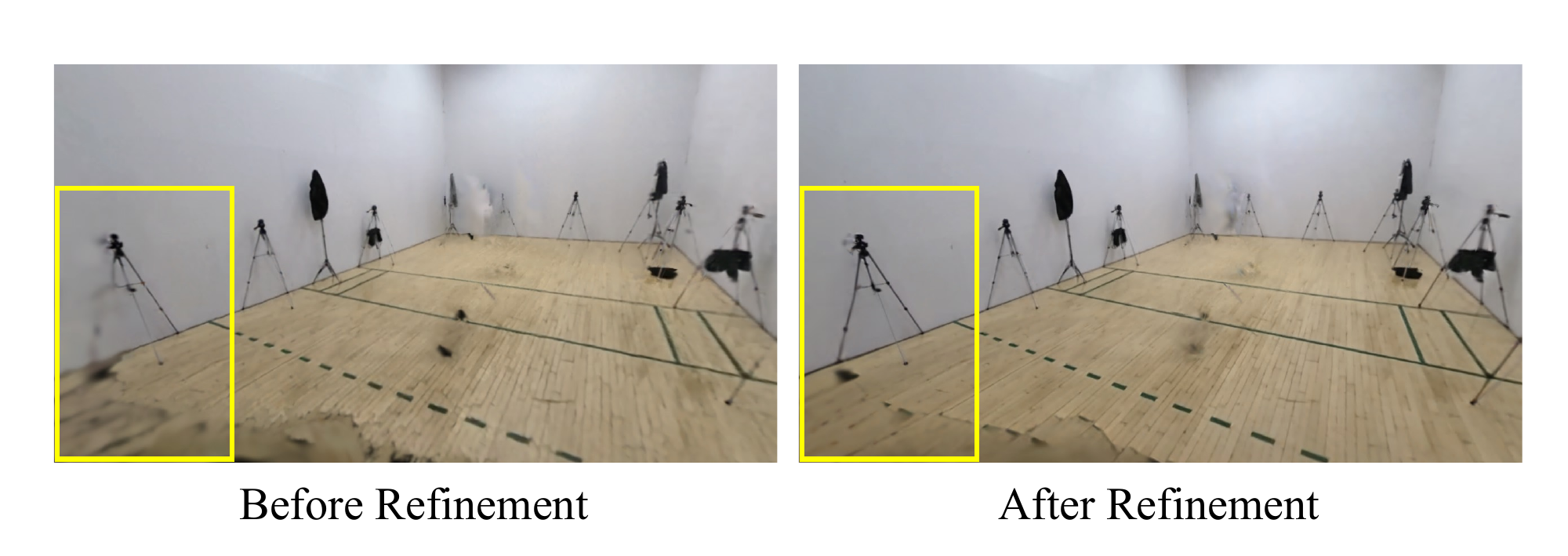}
    \caption{Effect of iterative refinement on background reconstruction (Sec.~\ref{sec:gaussian}). Before refinement (left), undersampled regions appear blurry. After refinement (right), clarity improves.}
    \Description{Before and after comparison of iterative refinement.}
    \label{fig:iterative_qual}
  \end{minipage}
  \hfill
  \begin{minipage}[b]{0.49\textwidth}
    \centering
    \includegraphics[width=0.95\linewidth]{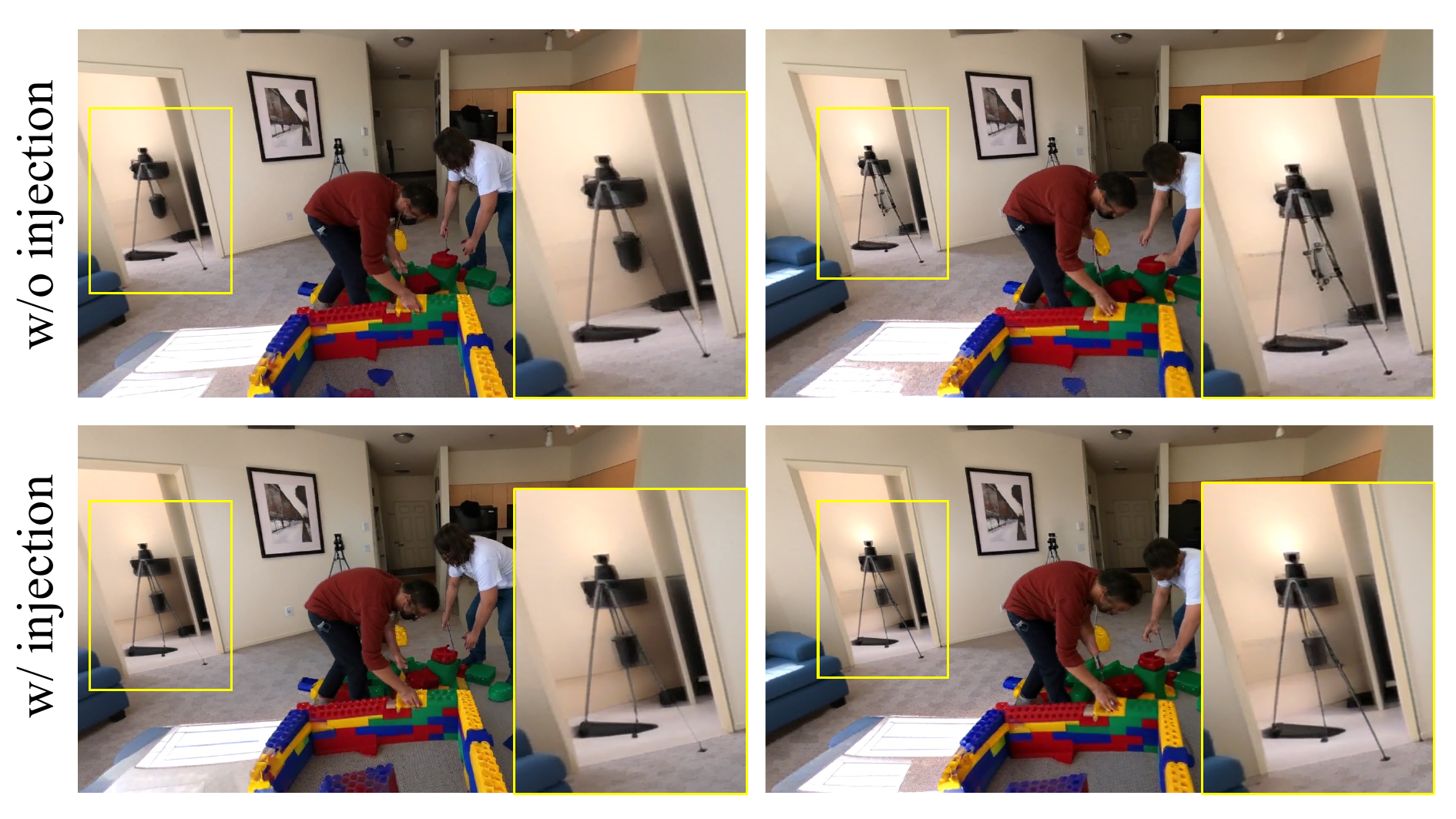}
    \caption{Ablation on motion-adaptive consistency injection (Sec.~\ref{sec:refinement}). Without injection (top), enhanced frames exhibit flickering. With injection (bottom), consistency improves.}
    \Description{Ablation on consistency injection.}
    \label{fig:ablation_inject}
  \end{minipage}
\end{figure*}

Motion-adaptive consistency injection reduces warp error by 23\%, improving frame-to-frame consistency. The gains in temporal coherence and per-frame sharpness come at the cost of a slight LPIPS increase, but this trade-off favors viewing experience as human perception is sensitive to flickering. Figure~\ref{fig:ablation_inject} shows that injection eliminates inconsistencies in static regions.

\begin{table}[t]
\centering
\caption{Seen vs.\ unseen fidelity on 6 360\degree{} scenes (Table~\ref{tab:360}).}
\label{tab:seen_unseen_main}
\begin{tabular}{l|cc|c}
\toprule
& PSNR$\uparrow$ & SSIM$\uparrow$ & Temporal L1$\downarrow$ \\
\midrule
Seen & \textbf{20.88} & \textbf{0.677} & \textbf{0.059} \\
Unseen & 18.55 & 0.556 & 0.178 \\
\bottomrule
\end{tabular}
\end{table}

\paragraph{Seen/Unseen Fidelity.}
To verify that generative priors preserve observed content, we classify each evaluation pixel as ``seen'' or ``unseen.'' For each evaluation pixel, we unproject it to 3D using estimated depth and project it into each training camera. If the projected depth matches the training camera's depth map within 5\% (i.e., the surface is visible and not occluded from that training view), the pixel is marked as seen (Table~\ref{tab:seen_unseen_main}).
The 2.3\,dB PSNR gap shows that seen regions maintain higher fidelity to ground truth than unseen ones. Temporal consistency (measured by optical-flow-warped L1 between consecutive frames) is also notably better for seen regions (0.059 vs.\ 0.178), as unseen regions lack direct observation and must be recovered by the generative prior.
Our sparse-to-dense view synthesis (Sec.~\ref{sec:view_synthesis}) conditions on input images and depths in 3D, making it reliable for seen regions while inpainting only unseen areas. The recursive enhancement module (Sec.~\ref{sec:refinement}) is conditioned on reference images from training views, preserving ground-truth appearance for seen regions. This confirms that our generative priors improve perceptual quality without compromising geometric fidelity in observed regions.

\begin{table}[t]
\centering
\caption{Sensitivity to input perturbation on 6 360\degree{} scenes (Table~\ref{tab:360}).}
\label{tab:robustness_main}
\begin{tabular}{l|cc}
\toprule
Config & PSNR$\uparrow$ & SSIM$\uparrow$ \\
\midrule
Full pipeline & \textbf{20.11} & \textbf{0.654} \\
\midrule
+ Mask erosion 5\,/\,10\,px & 20.04\,/\,20.03 & 0.653\,/\,0.653 \\
+ SMPL noise $\sigma$=3\degree\,/\,5\degree & 20.02\,/\,19.97 & 0.652\,/\,0.651 \\
+ 2D Keypoint noise $\sigma$=3\,/\,5\,px & 19.99\,/\,19.91 & 0.651\,/\,0.649 \\
\bottomrule
\end{tabular}
\end{table}

\paragraph{Sensitivity to Input Noise.}
Our pipeline relies on pre-trained models for segmentation, SMPL estimation, and keypoint detection. To test robustness to their errors, we inject realistic perturbations into each module output and re-train the full pipeline (Table~\ref{tab:robustness_main}).
We erode segmentation masks by 5 and 10\,px to simulate boundary imprecision. Since our pipeline applies 21 px mask dilation, the erosion is fully compensated (0.07--0.08\,dB drop).
We add Gaussian noise to SMPL joint angles ($\sigma$=3\degree, 5\degree) to simulate pose estimation errors from occlusion. The pipeline's joint optimization partially recovers from the noisy initialization, limiting degradation to 0.09--0.14\,dB PSNR.
We add pixel noise to 2D keypoints ($\sigma$=3, 5\,px at 1280$\times$704) before multi-view triangulation to simulate detection inaccuracy. Multi-view triangulation averages out per-view noise, limiting degradation to 0.12--0.20\,dB.


\subsection{Applications}

\begin{figure*}[p]
  \centering
  \begin{minipage}[b]{0.49\textwidth}
    \centering
    \includegraphics[width=0.95\linewidth]{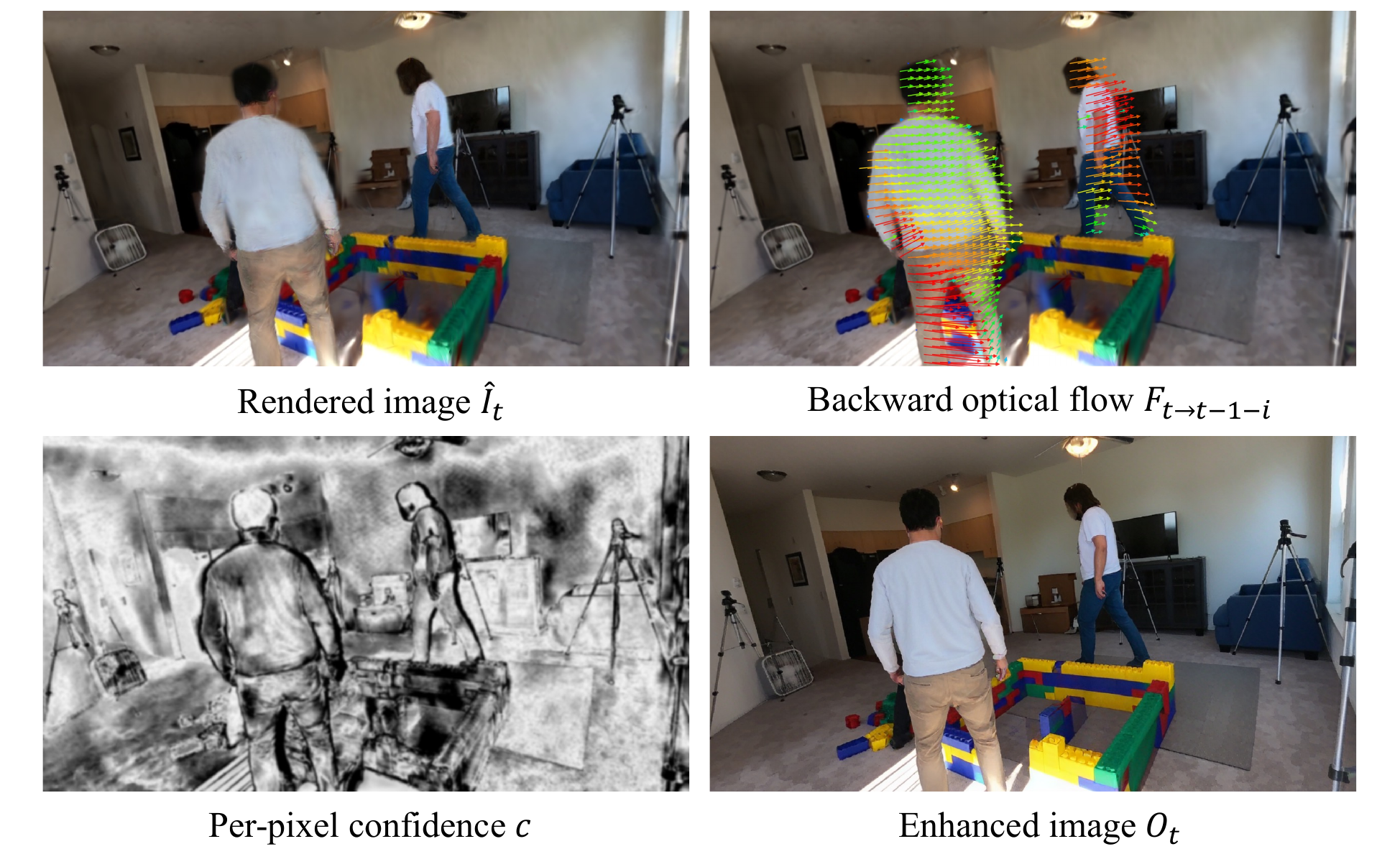}
    \caption{Visualization of motion-adaptive consistency injection (Sec.~\ref{sec:refinement}). From left to right, top to bottom: rendered input, backward optical flow, per-pixel confidence map, and enhanced output.}
    \Description{Visualization of EMA consistency injection.}
    \label{fig:flow_diagram}
  \end{minipage}
  \hfill
  \begin{minipage}[b]{0.49\textwidth}
    \centering
    \includegraphics[width=0.95\linewidth]{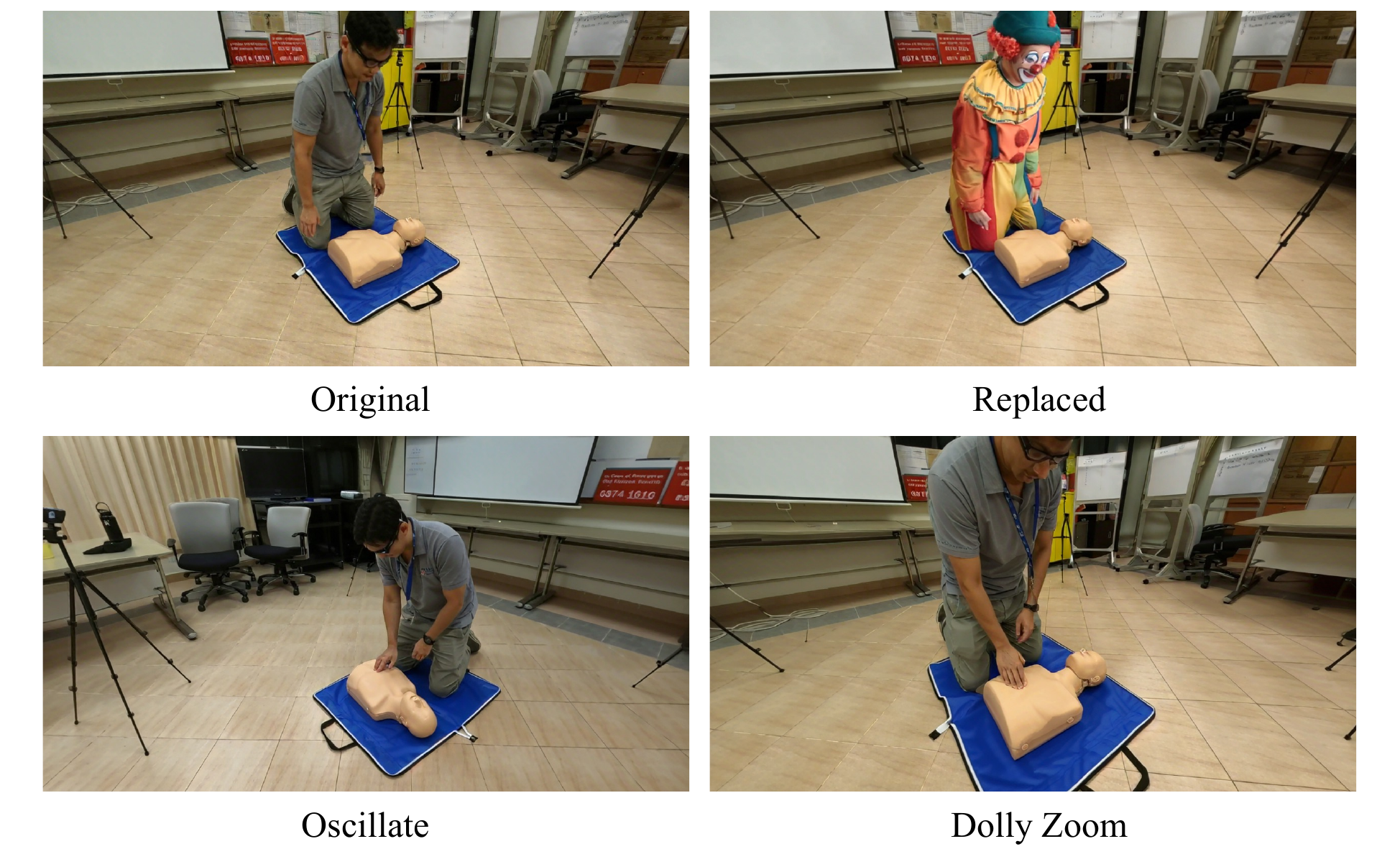}
    \caption{Applications. Top: human replacement with a different identity (right) while keeping the background (left). Bottom: novel trajectory rendering with oscillating motion (left) and dolly zoom (right).}
    \Description{Applications: human replacement and novel trajectory rendering.}
    \label{fig:application}
  \end{minipage}
\end{figure*}

\paragraph{Novel Camera Trajectories.}
The Gaussian representation supports rendering from arbitrary camera trajectories. Figure~\ref{fig:application} demonstrates two examples: \textit{dolly zoom} (translating toward the subject while widening FOV) and \textit{oscillating trajectory} (lateral periodic motion creating parallax), achieved by defining the desired camera path.
\paragraph{Human Replacement.}
Our decoupled pipeline with human parameters also enables actor replacement. Given a reference image, we generate a new human Gaussian using a feed-forward model~\cite{qiu2025lhm} and composite it into the existing background, transferring the original SMPL pose to the new identity. For rendering enhancement, we create a single reference image by editing the original frame to insert the new human. Since humans and backgrounds are reconstructed independently, this requires no modification to the scene (Figure~\ref{fig:application}). See our project page for video results.

\subsection{Limitations}

Our method has three limitations. (See Appendix~\ref{sec:appendix_limitations} for visual examples.)
First, fine details such as faces and hands remain challenging because high-frequency variations are difficult to capture from sparse views.
Second, our SMPL-based pipeline does not handle dynamic objects such as basketballs or props, leading to their disappearance in rendered outputs.
Third, shadows baked into the static background at $t{=}0$ do not follow human motion at later timesteps.
Incorporating hand-held object models and dynamic shadow synthesis are promising future directions.

\section{Conclusion}

We present \ours{}, a pipeline for high-fidelity 4D human-scene reconstruction for in-the-wild studio capture. Our key insight is that backgrounds and humans benefit from different priors.
Diffusion models synthesize dense supervision for backgrounds, while parametric body models constrain human geometry from limited views.
This decoupled approach enables high-quality reconstruction and applications such as novel camera trajectories and actor replacement.
Combined with temporally coherent diffusion enhancement, our method achieves state-of-the-art results across diverse real-world datasets, bringing high-fidelity 4D human-scene capture from sparse, low-overlap cameras closer to practical deployment.

\begin{acks}
This work was supported by IITP grants funded by the Korea government (MSIT) (RS-2021-II211343: AI Graduate School Program - Seoul National University (5\%), RS-2025-25442338: AI star Fellowship Support Program (45\%), and RS-2025-02303703: Real-world multi-space fusion and 6DoF free-viewpoint immersive visualization for extended reality (50\%))
\end{acks}

\clearpage

\begin{figure*}[p]
  \centering
  \includegraphics[width=0.9\textwidth]{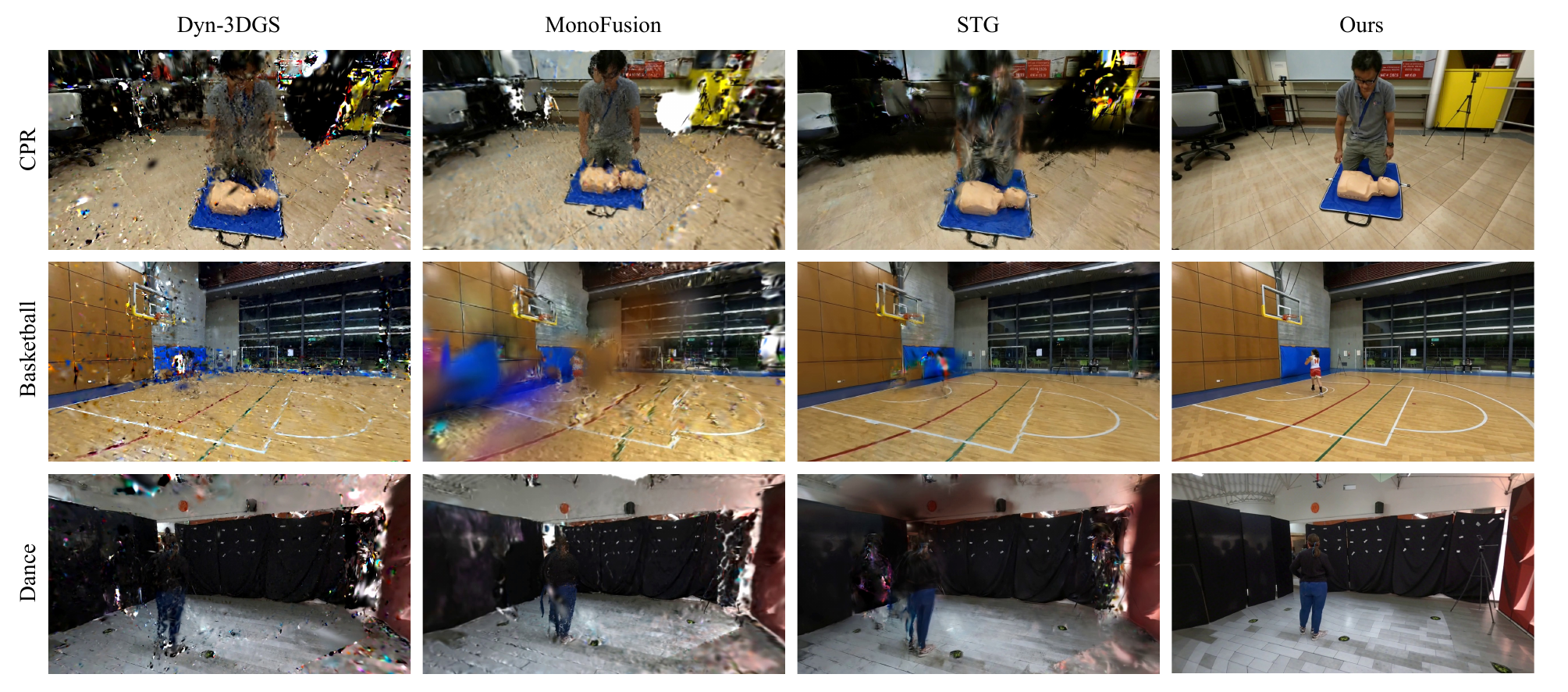}
  \caption{Qualitative results on EgoExo-4D~\cite{grauman2024ego}. Our method applies to diverse activities and environments.}
  \Description{Qualitative results on EgoExo-4D dataset.}
  \label{fig:qual_egoexo}
\end{figure*}

\clearpage
\bibliographystyle{ACM-Reference-Format}
\bibliography{main}

\clearpage
\appendix

\setcounter{figure}{0}
\setcounter{table}{0}
\setcounter{equation}{0}
\renewcommand{\thefigure}{S\arabic{figure}}
\renewcommand{\thetable}{S\arabic{table}}
\renewcommand{\theequation}{S\arabic{equation}}

\setcounter{topnumber}{3}
\setcounter{bottomnumber}{2}
\setcounter{totalnumber}{5}
\renewcommand{\topfraction}{0.9}
\renewcommand{\bottomfraction}{0.9}
\renewcommand{\textfraction}{0.05}
\renewcommand{\floatpagefraction}{0.6}

\raggedbottom

\section*{Appendix Overview}

Video results for the baseline comparisons and applications shown in the main paper and this appendix are available on our project page: \url{https://sisyphm.github.io/studiorecon-page/}.

This appendix provides additional implementation details and results for \ours{}. Appendix~\ref{sec:appendix_preprocessing} describes the preprocessing pipeline, including segmentation, depth estimation, and view synthesis. Appendix~\ref{sec:appendix_pose} details the human pose estimation process with cross-view association and SMPL fitting. Appendices~\ref{sec:appendix_background} and~\ref{sec:appendix_human} provide optimization configurations for background and human Gaussian reconstruction, respectively. Appendix~\ref{sec:appendix_refinement} explains the diffusion-based rendering enhancement. Appendix~\ref{sec:appendix_runtime} analyzes runtime and scalability. Appendix~\ref{sec:appendix_results} presents additional visualizations and evaluations. Appendix~\ref{sec:appendix_ablation} provides robustness analysis. Finally, Appendix~\ref{sec:appendix_limitations} discusses limitations.

\section{Preprocessing}
\label{sec:appendix_preprocessing}

We use $N=4$ input cameras in all experiments.
We employ SAM3~\cite{carion2025sam} with the text prompt ``person'' for actor segmentation.
Camera parameters and per-frame depth maps are estimated using Pi3~\cite{wang2025pi} with a confidence threshold of 0.1 and 21-pixel mask dilation to ensure complete actor coverage.
We synthesize $L=481$ novel views at $1280 \times 704$ resolution using GEN3C~\cite{ren2025gen3c} with guidance scale 1.0, where camera poses are interpolated via spherical linear interpolation (SLERP) for rotation and linear interpolation for translation.

\section{Human Pose Estimation}
\label{sec:appendix_pose}

Per-view actor detection and tracking are performed using CoMotion~\cite{newell2025comotion}, which provides SMPL pose and shape estimates, 2D keypoints, and within-view identity tracking.

Cross-view identity association employs the Hungarian algorithm on combined spatial and pose affinity at reference timestep $t_0 = 0$, with affinity weights $w_p = 0.9$ (spatial) and $w_\theta = 0.1$ (pose), and a distance threshold of 0.3m for valid matches. For each detected person, we estimate their 3D world position by unprojecting the pelvis location:
\begin{equation}
\mathbf{p}_{world} = R_n^{-1} \cdot \left( d \cdot K_n^{-1} \cdot \begin{bmatrix} u \\ v \\ 1 \end{bmatrix} - \mathbf{t}_n \right)
\end{equation}
where $(u, v)$ is the 2D pelvis location, $d$ is the depth value, and $(K_n, R_n, \mathbf{t}_n)$ are camera intrinsics and extrinsics. We then apply robust DLT triangulation with Huber loss to obtain 3D joint and vertex positions.
SMPL parameters are fitted through 3D-to-3D optimization over 30 iterations with a $\beta$ regularization coefficient of 10.0.
Scale alignment to the pipeline's coordinate system is performed via bone-length matching using L-BFGS optimization over 100 iterations.
Temporal consistency of the fitted pose parameters is enforced through bidirectional exponential smoothing ($\alpha = 0.5$) followed by Gaussian filtering ($\sigma = 1$).

\section{Background Reconstruction}
\label{sec:appendix_background}

Background Gaussians are optimized for 7,000 iterations over the $L$ synthesized views with spherical-harmonic degree 3.
The loss function combines L1 reconstruction ($\lambda = 0.7$), SSIM ($\lambda = 0.3$), LPIPS ($\lambda = 0.1$), and density regularization ($\lambda = 0.1$).
Position learning rate decays from $1.6 \times 10^{-4}$ to $1.6 \times 10^{-6}$.
Densification is performed every 500 iterations up to iteration 3,000, with a gradient threshold of 0.0002.

\paragraph{Ground-Truth View Weighting.}
Views near original camera positions receive up to $3\times$ higher loss weight via cosine falloff, prioritizing ground-truth supervision over interpolated views:
\begin{equation}
w_i = 1 + \lambda_{gt} \cdot \frac{1}{2}\left(1 + \cos\left(\frac{\pi \cdot d_i}{d_{max}}\right)\right)
\end{equation}
where $d_i$ is the angular distance to the nearest original camera, $d_{max}$ is the cutoff distance, and $\lambda_{gt} = 2$ controls the boost magnitude.

\paragraph{Iterative Refinement.}
At iteration 3,500, we create 240 additional views with $\pm$15\% height oscillation relative to the scene radius, which are refined using Difix3D+~\cite{wu2025difix3d+} and incorporated with a reduced loss weight of 0.7.

\section{Human Reconstruction}
\label{sec:appendix_human}

Human Gaussians are optimized for 10,000 iterations on the 4 input videos with spherical harmonics degree 2.
The loss function combines multiple terms:
\begin{equation}
\mathcal{L}_{hn} = \mathcal{L}_1 + \lambda_s\mathcal{L}_{SSIM} + \lambda_l\mathcal{L}_{LPIPS} + \lambda_m\mathcal{L}_{mask} + \lambda_d\mathcal{L}_{den} + \lambda_p\mathcal{L}_{pose}
\end{equation}
where L1 ($\lambda = 0.8$) and SSIM ($\lambda = 0.2$) measure photometric reconstruction, LPIPS ($\lambda = 0.1$) provides perceptual supervision, mask loss ($\lambda = 0.01$) supervises the rendered silhouette, density regularization ($\lambda = 0.1$) prevents spurious Gaussians, and pose regularization ($\lambda = 50.0$) enforces temporal smoothness of SMPL parameters. We jointly optimize Gaussian attributes and SMPL pose parameters.
Position learning rate decays from $1.0 \times 10^{-3}$ to $2.0 \times 10^{-6}$.
Densification terminates at iteration 1,500, as actors are initialized from SMPL mesh vertices.

\paragraph{Linear Blend Skinning.}
We employ Linear Blend Skinning with SMPL weights computed from a $ 64^3$-voxel grid.

\paragraph{Temporal Deformation Network.}
Temporal deformation is enabled at iteration 3,000 using an 8-layer MLP with 256 hidden units and a skip connection at layer 4. The network predicts per-Gaussian residuals:
\begin{equation}
(\Delta\mathbf{c}, \Delta\alpha) = f_\theta(\gamma(\boldsymbol{\mu}_c), \gamma(t))
\end{equation}
where $\gamma(\cdot)$ denotes positional encoding with multires=10, yielding 63D for spatial coordinates $\boldsymbol{\mu}_c$ and 21D for normalized time $t$. The residuals are added to the base Gaussian attributes before rendering.

\section{Recursive Enhancement Module}
\label{sec:appendix_refinement}

\paragraph{Consistency Injection.}
At inference time, we employ per-frame reference mode with single-step denoising at timestep $t=200$.
Temporal consistency is achieved through motion-adaptive EMA input injection over 3 previous frames with $\alpha = 0.3$ and decay factor 0.5.
Injection strength is determined by warp quality: we compute the per-pixel RGB error between the warped previous output and the current input, with an error threshold $\tau_e = 30$ (on a 0--255 scale).
Pixels with low warp error receive full injection, while misaligned regions (due to motion or occlusion) receive minimal injection.
A Gaussian blur with a kernel size of 11 smooths the confidence map, reducing sharp transitions.

\section{Runtime Analysis}
\label{sec:appendix_runtime}

Table~\ref{tab:runtime} reports the per-stage runtime breakdown on a single NVIDIA A6000 GPU.

\begin{table}[tbp]
\centering
\caption{Per-stage runtime breakdown on a single NVIDIA A6000 GPU.}
\label{tab:runtime}
\begin{tabular}{l|c}
\toprule
Stage & Time \\
\midrule
View Synthesis (GEN3C) & 50 min \\
Human Pose Estimation & 3 min \\
Background Reconstruction & 30 min \\
Human Reconstruction & 40 min \\
Recursive Enhancement & 3.3 s/frame \\
\bottomrule
\end{tabular}
\end{table}

\paragraph{Bottleneck.}
View synthesis using GEN3C~\cite{ren2025gen3c} constitutes the primary bottleneck due to the computational cost of video diffusion models. However, this stage is easily parallelized across multiple GPUs, as each trajectory segment can be processed independently. With 8 GPUs, the view synthesis time reduces to approximately 13 minutes.

\paragraph{Runtime Scaling.}
Human reconstruction time scales linearly with the number of actors (approximately 3 minutes per additional human), while other stages remain largely unaffected. Background reconstruction and human reconstruction can run in parallel once preprocessing completes.

\paragraph{Parallelization.}
Multi-GPU processing is supported for the preprocessing and view synthesis stages. Additionally, most preprocessing stages (segmentation, depth estimation, and pose estimation) can execute concurrently. Our pipeline is designed for offline processing, where diffusion-based rendering enhancement adds per-frame computation for the final output.

\paragraph{End-to-End Time.}
For a typical scene with 121 frames and 1-3 actors, the total time to produce a final rendered video is approximately 130 minutes on a single GPU, or 60 minutes with 8 GPUs.

\section{Additional Results}
\label{sec:appendix_results}

\paragraph{Camera Configuration.}
To quantify the difficulty of our setting across EgoHumans, Harmony4D, Mobile Stage, and SelfCap, we compare against the camera selection protocol of Kong~\textit{et al.}~\cite{kong2025generative}, a recent sparse-view method (Table~\ref{tab:camera_overlap}). Even with maximized spread, neighboring cameras in prior work still overlap substantially (28\degree). In our setting, each view observes a different part of the scene with heavy self-occlusion (70\degree).

\begin{table}[tbp]
\centering
\caption{Camera configuration comparison with prior sparse-view work.}
\label{tab:camera_overlap}
\begin{tabular}{l|cc}
\toprule
& Kong~\textit{et al.} & Ours \\
\midrule
Training cameras & 3 & 4 \\
Position angle from center & 17\degree & 117\degree \\
Nearest-neighbor viewing angle & 28\degree & 70\degree \\
\bottomrule
\end{tabular}
\end{table}

\begin{figure}[tbp]
  \centering
  \includegraphics[trim={0 3cm 0 0}, clip, width=0.6\linewidth]{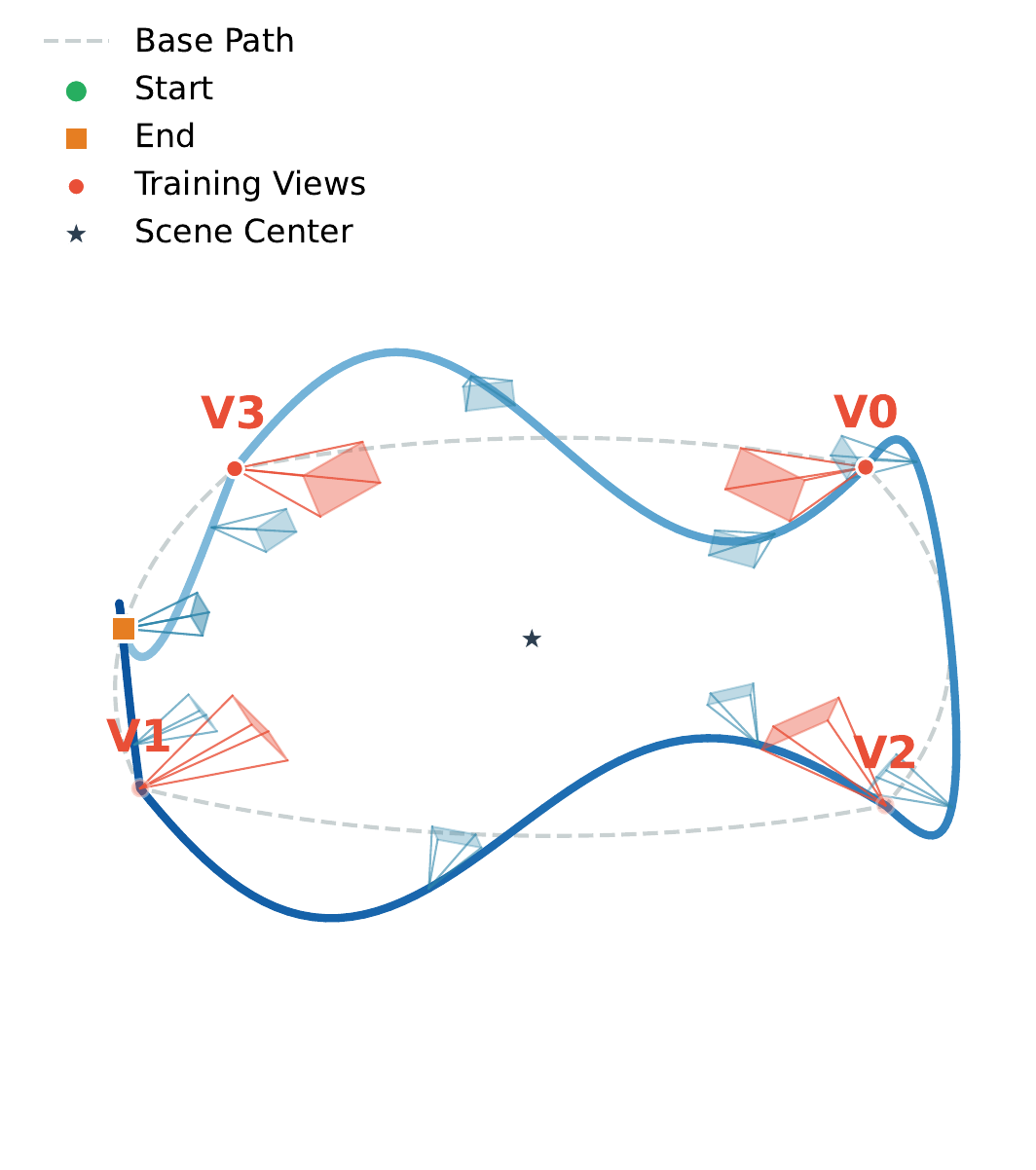}

  \caption{Augmented camera trajectory for iterative refinement. We interpolate between synthesized camera poses while adding sinusoidal height variation and compensating pitch rotation, providing additional supervision from elevated and lowered vantage points.}
  \Description{Visualization of augmented camera trajectory.}
  \label{fig:iterative_trajectory}
\end{figure}

\paragraph{Augmented Camera Trajectory.}
Figure~\ref{fig:iterative_trajectory} visualizes the augmented camera trajectory used for iterative refinement during background reconstruction. Starting from the 481 synthesized camera poses, we create 240 additional viewpoints by interpolating positions while adding sinusoidal height variation ($\pm$15\% of scene radius) and compensating pitch rotation. This provides supervision from elevated and lowered vantage points, improving reconstruction quality in undersampled regions such as floors and ceilings.

\begin{figure}[tbp]
  \centering
  \includegraphics[trim={0cm 0 0cm 0}, clip, width=\linewidth]{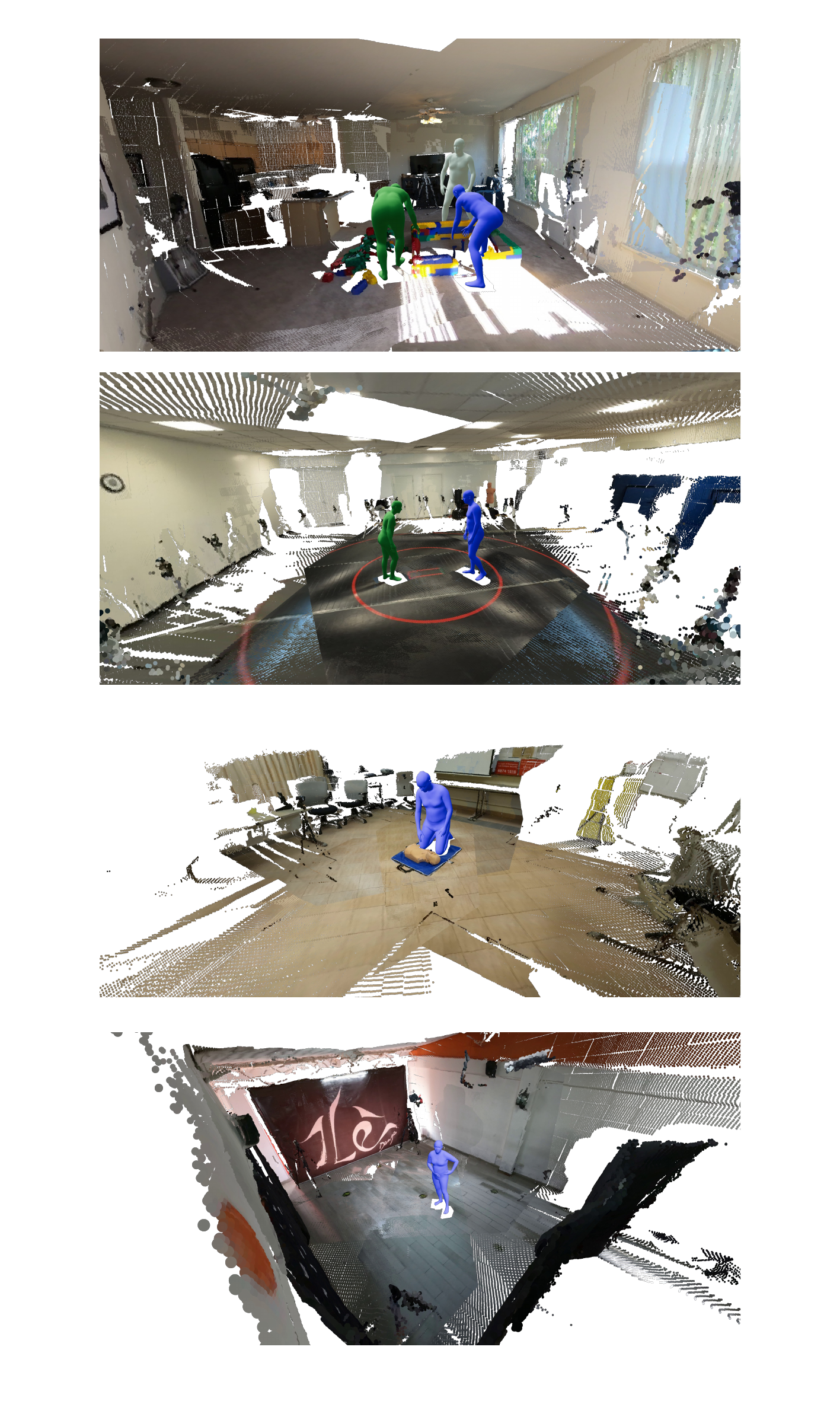}

  \caption{Initialized human pose visualization across four scenes. Point clouds are rendered with overlaid SMPL meshes, demonstrating accurate body pose estimation from sparse multi-view inputs.}
  \Description{Visualization of initialized human poses with SMPL meshes.}
  \label{fig:pose_visualization}
\end{figure}

\paragraph{Human Pose Estimation.}
Figure~\ref{fig:pose_visualization} shows the initialized human poses from our multi-view SMPL fitting pipeline, visualized as point clouds with overlaid SMPL meshes. The accurate alignment between the fitted body models and the scene geometry demonstrates the effectiveness of our cross-view association and 3D pose estimation.

\begin{figure}[tbp]
  \centering
  \includegraphics[width=\linewidth]{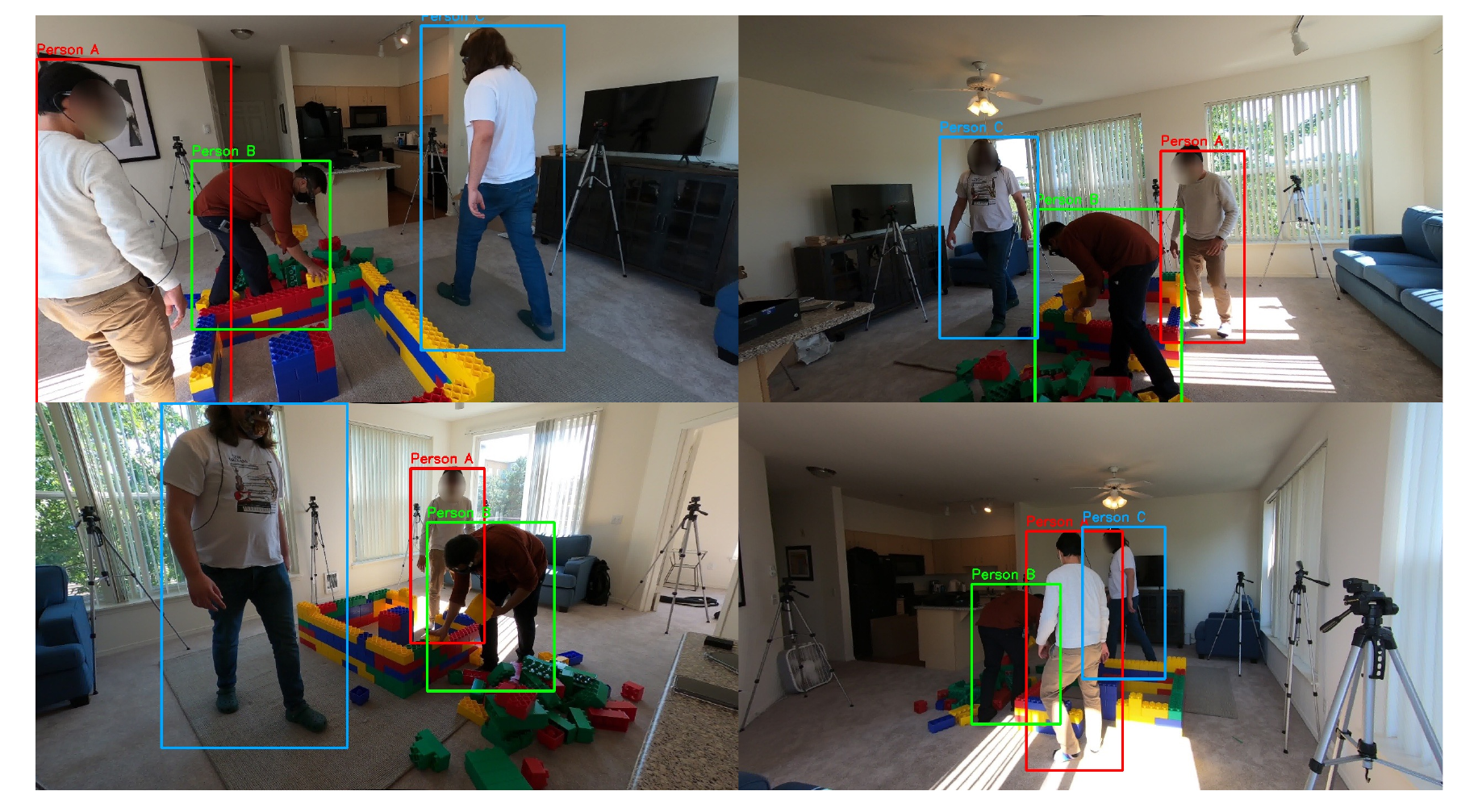}
  \caption{Cross-view identity association. Each color represents the same person matched across 4 cameras with low overlap. Our method achieves 97.8\% association accuracy across all 8 scenes from EgoHumans, Harmony4D, Mobile Stage, and SelfCap (Table~\ref{tab:ablation_association}).}
  \Description{Cross-view identity association visualization.}
  \label{fig:crossview}
\end{figure}

\paragraph{Cross-View Identity Association.}
In our low-overlap setting, cameras observe largely disjoint regions, making appearance-based matching infeasible. We instead lift each detection to 3D using estimated depth, and compute a pairwise affinity combining spatial proximity and SMPL pose similarity. The Hungarian algorithm assigns cross-view correspondences based on this affinity. Figure~\ref{fig:crossview} visualizes the results, where consistent bounding box colors indicate the same person matched across 4 cameras.

\begin{figure}[tbp]
  \centering
  \includegraphics[width=\linewidth]{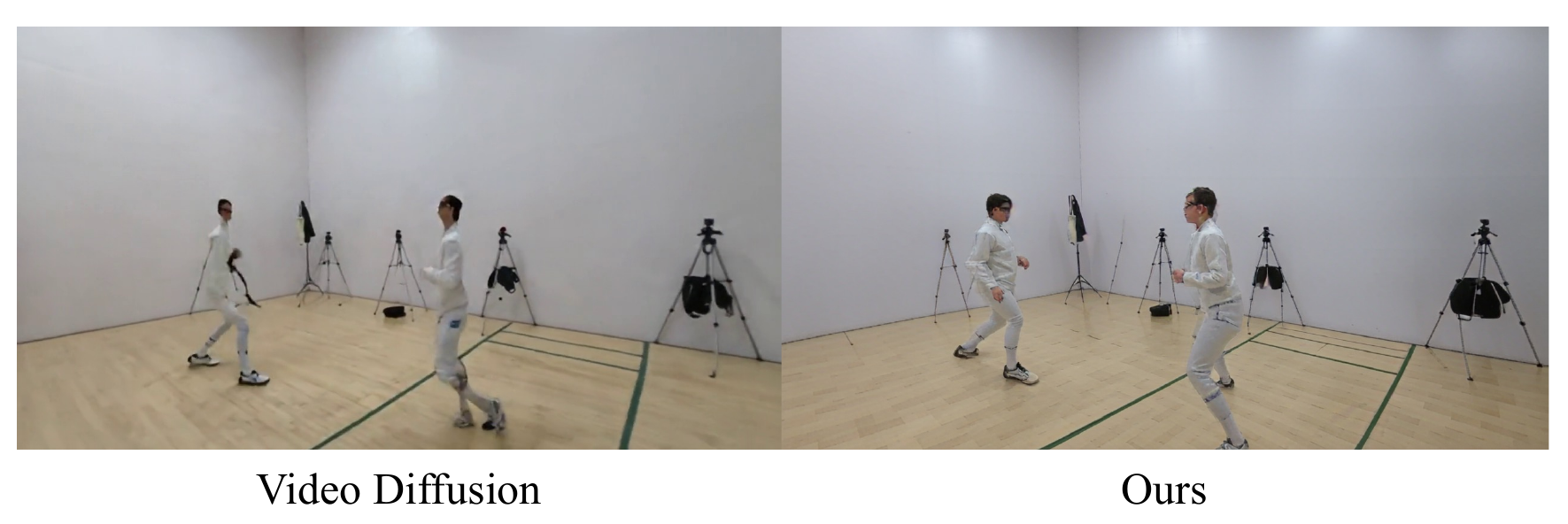}
  \caption{Video diffusion models~\cite{ren2025gen3c} produce geometrically inconsistent humans (left), while our explicit reconstruction maintains accurate body shape (right).}
  \Description{Comparison between video diffusion output and our reconstruction.}
  \label{fig:gen3c_comparison}
\end{figure}

\paragraph{Video Diffusion for Human Novel View Synthesis.}
Camera-controlled video diffusion models are designed for general novel view synthesis but degrade on dynamic humans. We validate this by comparing 2D pose accuracy between GEN3C~\cite{ren2025gen3c} outputs and our Gaussian reconstruction. Since GEN3C is not designed for multi-view video, we run it at key timesteps with 4 input views to perform static scene novel view synthesis. Using DWPose~\cite{yang2023effective} on 8 scenes across all four datasets (EgoHumans, Harmony4D, Mobile Stage, SelfCap) with 46 person instances, we compute PCK (Percentage of Correct Keypoints) against ground truth frames. As shown in Table~\ref{tab:pck} and Figure~\ref{fig:gen3c_comparison}, GEN3C produces geometrically inconsistent humans with significantly lower pose accuracy, while our method maintains accurate body geometry.

\begin{table}[tbp]
\centering
\caption{Pose accuracy of novel view synthesis methods on human subjects.}
\label{tab:pck}
\begin{tabular}{l|cc|c}
\toprule
Method & PCK@0.05$\uparrow$ & PCK@0.1$\uparrow$ & Detection Rate$\uparrow$ \\
\midrule
GEN3C & 55.6\% & 71.4\% & 97.6\% \\
Ours & \textbf{93.1\%} & \textbf{97.7\%} & \textbf{100\%} \\
\bottomrule
\end{tabular}

\end{table}

\paragraph{Enhancement on Baseline Methods.}
Our recursive enhancement module is designed to be generally applicable. To evaluate its effectiveness beyond our pipeline, we apply it to baseline methods (Dyn-3DGS~\cite{luiten2024dynamic}, MonoFusion~\cite{wang2025monofusion}, STG~\cite{li2024spacetime}) by using their rendered outputs as input to Difix3D+~\cite{wu2025difix3d+} with consistency injection. Table~\ref{tab:enhance_baselines} reports the results averaged across all 8 scenes from EgoHumans, Harmony4D, Mobile Stage, and SelfCap.

\begin{table*}[tbp]
\centering
\caption{Effect of applying our recursive enhancement module to baseline methods. All metrics averaged across all 8 scenes from EgoHumans, Harmony4D, Mobile Stage, and SelfCap.}
\label{tab:enhance_baselines}
\begin{tabular}{l|ccc|ccc}
\toprule
& \multicolumn{3}{c|}{Original} & \multicolumn{3}{c}{+ Enhancement (Ours)} \\
Method & PSNR$\uparrow$ & SSIM$\uparrow$ & LPIPS$\downarrow$ & PSNR$\uparrow$ & SSIM$\uparrow$ & LPIPS$\downarrow$ \\
\midrule
Dyn-3DGS & 16.68 & 0.441 & 0.563 & 17.66 \textcolor{green!50!black}{\scriptsize(+0.98)} & 0.522 \textcolor{green!50!black}{\scriptsize(+.081)} & 0.396 \textcolor{green!50!black}{\scriptsize(-.167)} \\
MonoFusion & 17.06 & 0.435 & 0.552 & 17.68 \textcolor{green!50!black}{\scriptsize(+0.62)} & 0.480 \textcolor{green!50!black}{\scriptsize(+.045)} & 0.383 \textcolor{green!50!black}{\scriptsize(-.169)} \\
STG & 17.21 & 0.522 & 0.477 & 17.72 \textcolor{green!50!black}{\scriptsize(+0.51)} & 0.542 \textcolor{green!50!black}{\scriptsize(+.020)} & 0.380 \textcolor{green!50!black}{\scriptsize(-.097)} \\
\bottomrule
\end{tabular}
\end{table*}

The enhancement improves perceptual quality across all baselines (LPIPS reduces by 20--31\%), demonstrating that single-step diffusion can effectively suppress common rendering artifacts such as floaters and blur. However, the gains in reconstruction fidelity (PSNR, SSIM) are modest, and the enhanced baselines remain substantially below our full pipeline (PSNR 20.44, LPIPS 0.198). This reflects an inherent limitation: diffusion-based enhancement cannot recover geometric information lost during reconstruction. When the underlying 3D representation is degraded due to insufficient viewpoint coverage, enhancement can only improve surface appearance, not correct structural errors.

\paragraph{Baselines with Dense GEN3C Supervision.}
To test whether baselines benefit from dense video diffusion supervision, we synthesized GEN3C views at 30 timesteps (481$\times$30 images, 30$\times$ our diffusion budget) and trained baselines on these. Table~\ref{tab:gen3c_baselines} reports full-image and foreground metrics across 3 EgoHumans scenes (Legoassemble, Tennis, Fencing).

\begin{table}[tbp]
\centering
\caption{Baselines trained with dense GEN3C supervision (481$\times$30 views) vs.\ ours using GEN3C only at $t{=}0$.}
\label{tab:gen3c_baselines}
\begin{tabular}{l|ccc|c}
\toprule
Method & PSNR$\uparrow$ & SSIM$\uparrow$ & LPIPS$\downarrow$ & FG PSNR$\uparrow$ \\
\midrule
Dyn-3DGS (4-view) & 16.57 & 0.421 & 0.486 & - \\
Dyn-3DGS+GEN3C & 17.05 & 0.581 & 0.549 & 15.09 \\
STG (4-view) & 17.83 & 0.516 & 0.426 & - \\
STG+GEN3C & 18.72 & 0.630 & 0.377 & 17.14 \\
\textbf{Ours w/o enh.} & \textbf{20.39} & \textbf{0.657} & \textbf{0.286} & \textbf{18.08} \\
\bottomrule
\end{tabular}
\end{table}

Dense GEN3C supervision largely improves baselines through better background coverage, but for foreground humans, ours achieves 18.08 vs.\ 17.14 FG PSNR with STG+GEN3C despite 30$\times$ less diffusion. GEN3C-supervised humans also exhibit pose inconsistency (Table~\ref{tab:pck}), confirming dense diffusion alone cannot replace SMPL-guided geometric constraints for dynamic regions.

\paragraph{GEN3C vs.\ Ours (Rendering Quality).}
Since GEN3C cannot directly consume multi-view video, we run it independently at each timestep. Across all 8 scenes from EgoHumans, Harmony4D, Mobile Stage, and SelfCap, our rendering on held-out evaluation cameras (PSNR/\allowbreak SSIM/\allowbreak LPIPS: 20.13/\allowbreak 0.656/\allowbreak 0.215) surpasses GEN3C's raw synthesized views (19.05/\allowbreak 0.612/\allowbreak 0.329) despite using GEN3C only at $t{=}0$, as our decoupled pipeline avoids propagating GEN3C's geometric inconsistencies into dynamic regions (Figure~\ref{fig:gen3c_comparison}).

\section{Robustness Analysis}
\label{sec:appendix_ablation}

We analyze the robustness of our pipeline by disabling individual components and measuring the impact on reconstruction quality. We evaluate across all 6 scenes from Table~\ref{tab:360} (Legoassemble, Tennis, Fencing from EgoHumans; Sword, Karate, Grappling from Harmony4D). All qualitative comparisons in this section are shown without the recursive enhancement module for fair evaluation.

\begin{figure}[H]
  \centering
  \includegraphics[width=\linewidth]{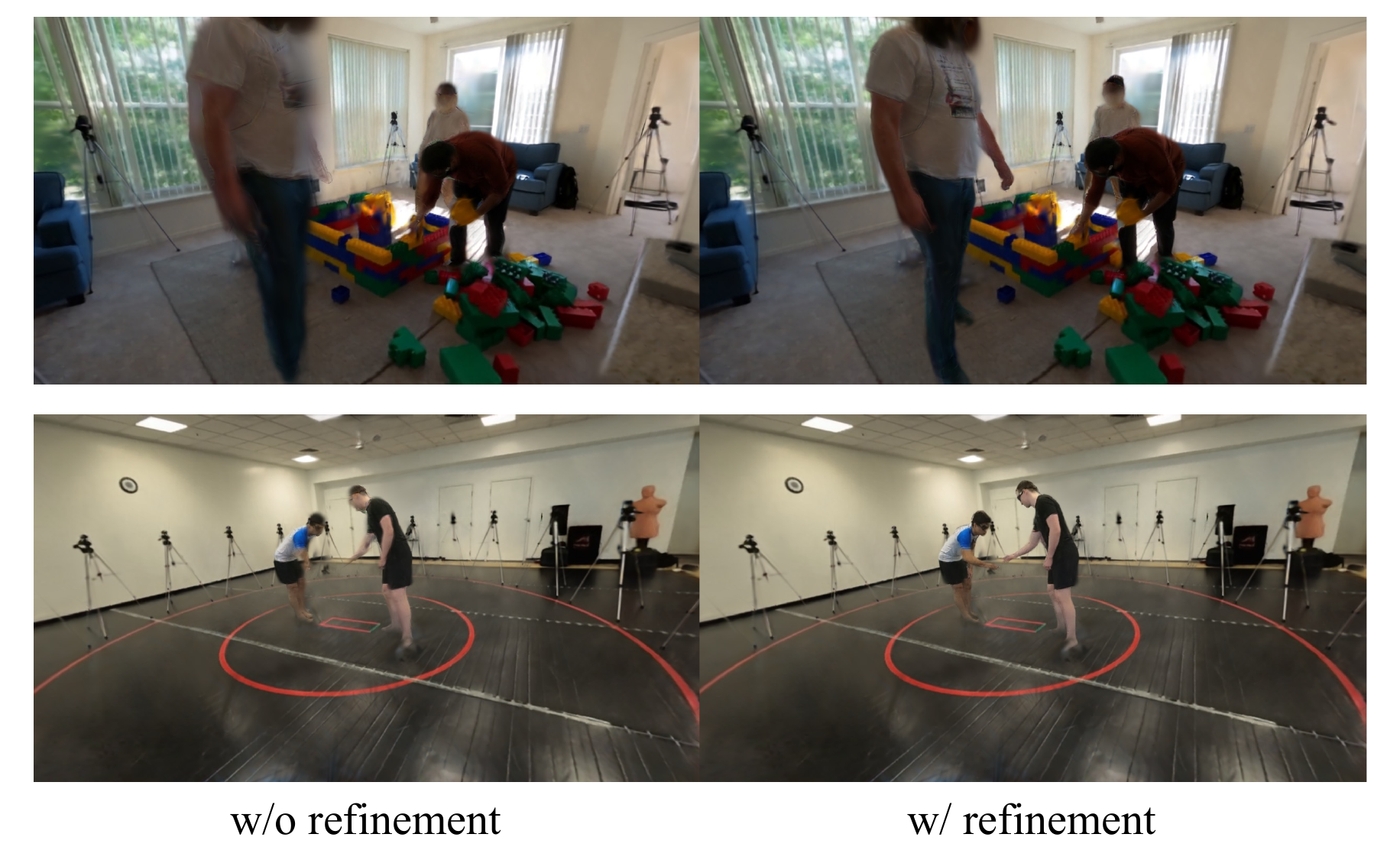}
  \caption{Effect of multi-view pose refinement. Without refinement (left), inaccurate SMPL poses cause blurry or semi-transparent body parts. With refinement (right), poses are corrected via cross-view triangulation, producing a sharper human reconstruction.}
  \Description{Qualitative comparison with and without multi-view pose refinement.}
  \label{fig:sensitivity_nlf}
\end{figure}

\paragraph{Multi-View Pose Refinement.}
Our multi-view human pose estimation (Sec.~\ref{sec:human_init}) refines initial per-view SMPL estimates through cross-view triangulation and 3D-to-3D SMPL fitting. We ablate this by using the initial fused per-view SMPL directly without multi-view refinement. As shown in Figure~\ref{fig:sensitivity_nlf}, less accurate poses cause Gaussians to misalign with the actual human regions, resulting in blurry or semi-transparent body parts in the rendered output.

\paragraph{Temporal Smoothing.}
We enforce temporal consistency of SMPL parameters through bidirectional exponential smoothing and Gaussian filtering. Table~\ref{tab:smoothing_jitter} shows that removing both significantly increases pose jitter, which appears as flickering in rendered video.

\begin{table}[tbp]
\centering
\caption{Effect of temporal smoothing on SMPL pose stability across 6 360\degree{} scenes (Table~\ref{tab:360}). Jitter is measured as the mean joint acceleration across all actors and frames.}
\label{tab:smoothing_jitter}
\begin{tabular}{l|cc}
\toprule
& Translation (mm/f$^2$)$\downarrow$ & Pose ($^\circ$/f$^2$)$\downarrow$ \\
\midrule
w/o smoothing & 2.74 & 40.67 \\
w/ smoothing & \textbf{1.29} & \textbf{13.95} \\
\bottomrule
\end{tabular}
\end{table}

\begin{figure}[H]
  \centering
  \includegraphics[width=\linewidth]{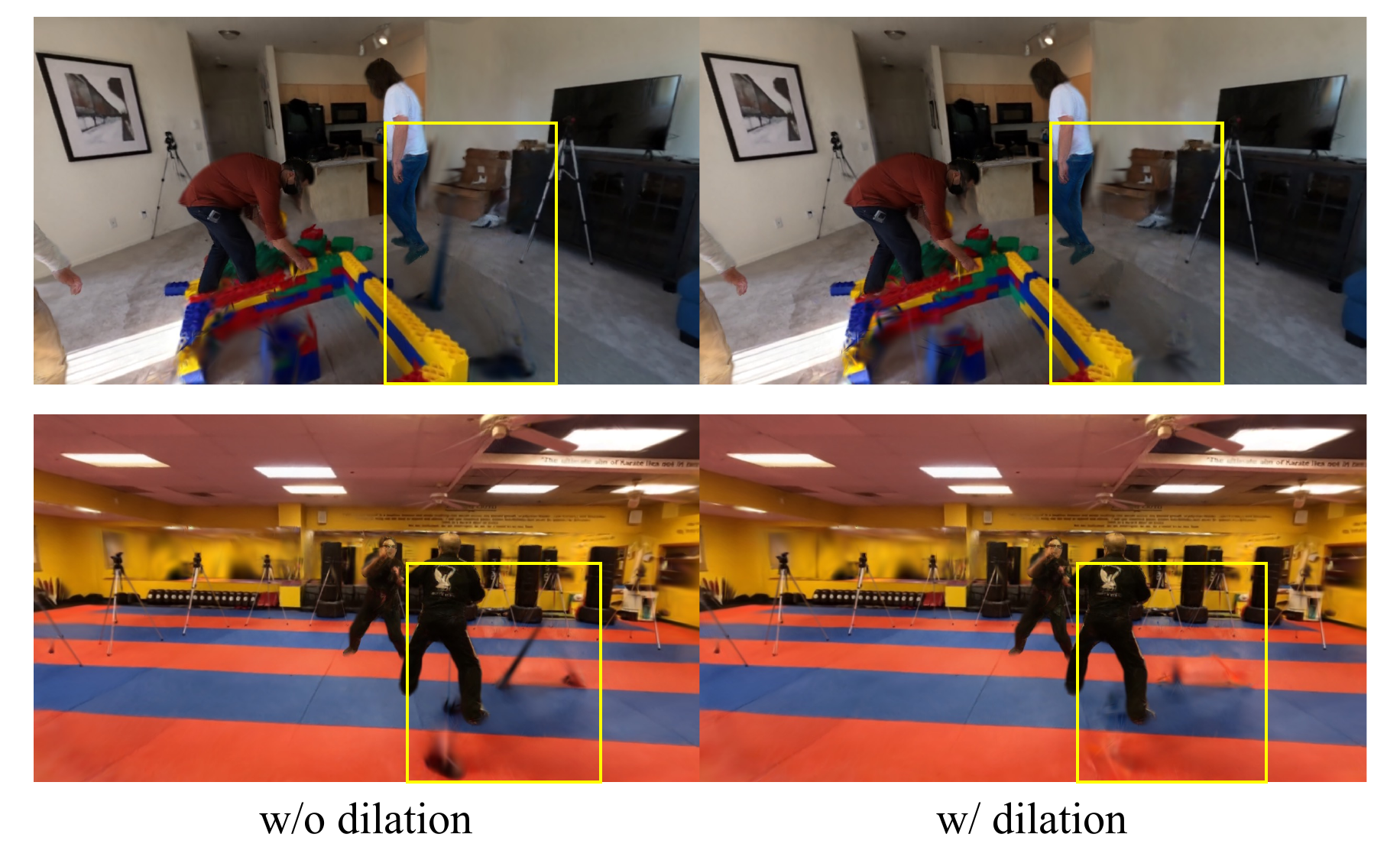}
  \caption{Effect of mask dilation. Without dilation (left), insufficiently masked human regions at $t{=}0$ are baked into the static background Gaussians, leaving ghosting artifacts as humans move. With 21px dilation (right), the background is cleanly separated.}
  \Description{Qualitative comparison with and without mask dilation.}
  \label{fig:sensitivity_mask}
\end{figure}

\paragraph{Mask Dilation.}
During background reconstruction, we dilate actor masks by 21 pixels to ensure human regions are fully excluded. Without dilation, tight masks leave residual human pixels in the background Gaussians, causing ghosting artifacts near human boundaries (Figure~\ref{fig:sensitivity_mask}). With dilation, the masked regions are under-supervised and may contain minor artifacts, but these are within the range that our recursive enhancement module can effectively correct.

\section{Limitations}
\label{sec:appendix_limitations}

\begin{figure}[H]
  \centering
  \includegraphics[width=\linewidth]{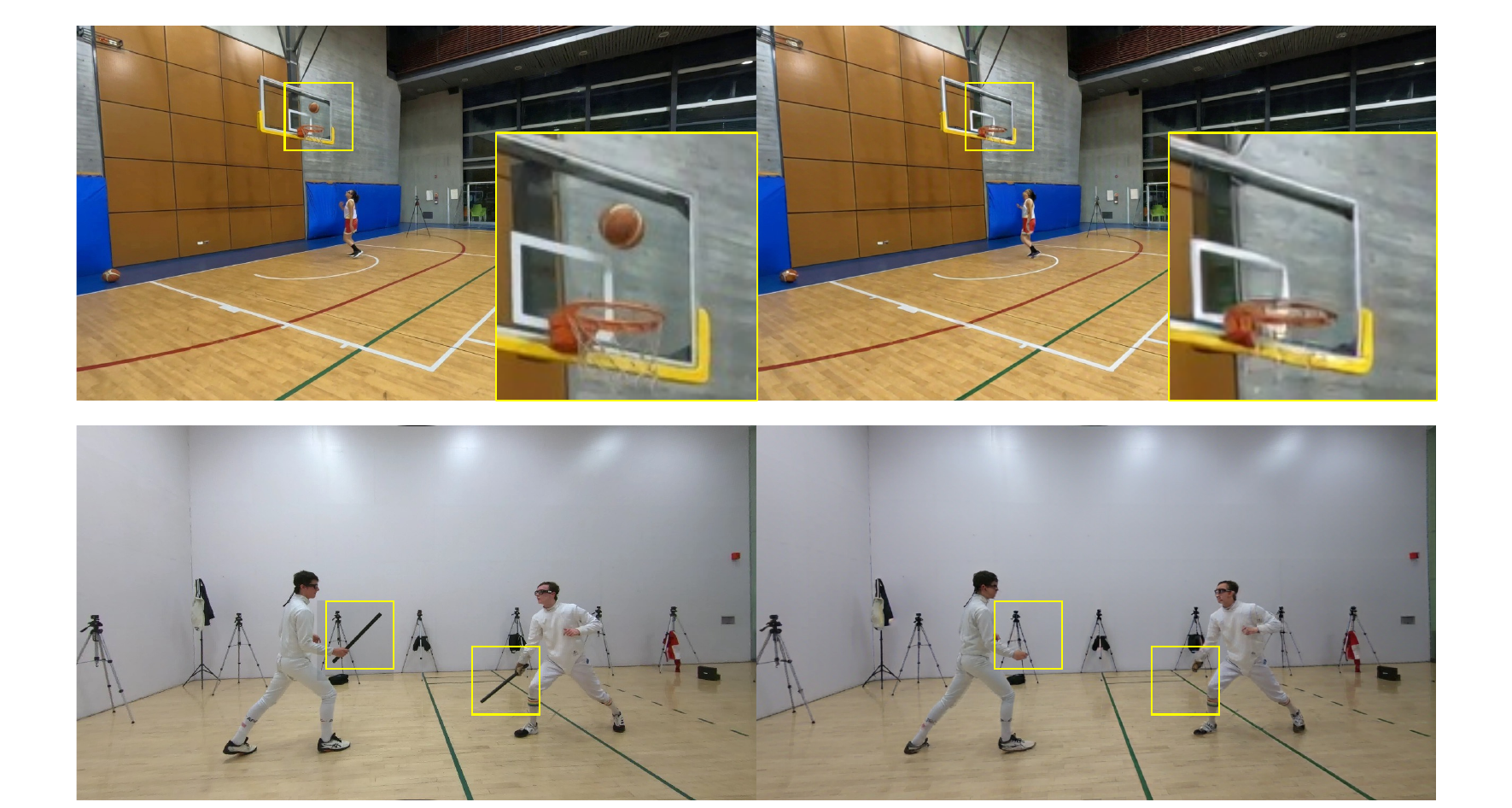}
  \caption{Limitations of our method. Comparing ground truth (left) with our reconstruction (right), dynamic objects held by actors are not reconstructed because they lie outside the SMPL body model.}
  \Description{Comparison showing limitations of our method.}
  \label{fig:limitations}
\end{figure}

\begin{figure}[H]
  \centering
  \includegraphics[width=\linewidth]{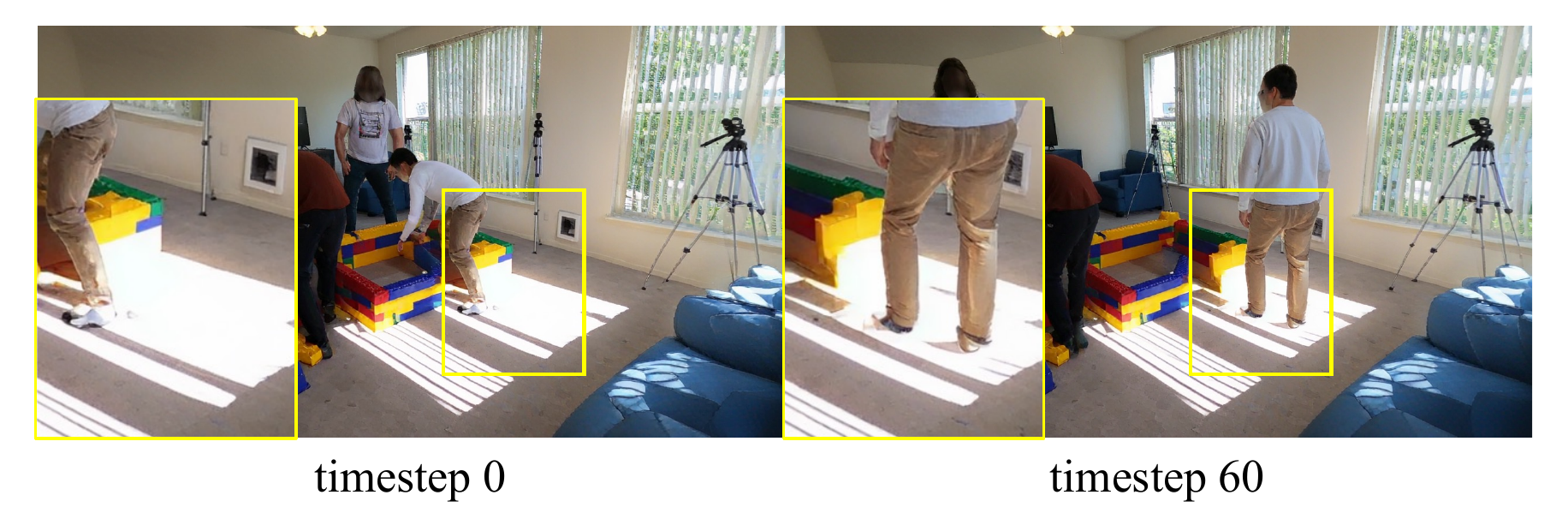}
  \caption{Shadow artifacts. Shadows baked into the static background at $t{=}0$ (left) remain fixed and do not follow human motion at later timesteps (right).}
  \Description{Shadow artifact example.}
  \label{fig:shadow}
\end{figure}

As shown in Figure~\ref{fig:limitations}, dynamic objects manipulated by actors are not reconstructed. This stems from our reliance on the SMPL body model, which represents only the human body surface and does not model held objects or accessories. Additionally, since shadows are baked into the static background at $t{=}0$, they remain fixed and do not follow human motion, resulting in incorrect shadows as humans move (Figure~\ref{fig:shadow}).

\end{document}